\documentclass[10pt,twocolumn,letterpaper]{article}

\usepackage{iccv}
\usepackage{times}
\usepackage{epsfig}
\usepackage{graphicx}
\usepackage{amsmath}
\usepackage{amssymb}
\usepackage{algorithm,algorithmic}
\usepackage{amsmath}
\usepackage{titles}
\usepackage{beamerarticle}
\usepackage{subcaption}

\usepackage[T1]{fontenc}
\usepackage[utf8]{inputenc}
\usepackage{authblk}

\usepackage{color}
\definecolor{orange}{rgb}{0.8, 0.6, 0.2}
\definecolor{teal}{rgb}{0.0, 0.4, 0.4}
\definecolor{purple}{rgb}{0.65,0,0.65}
\definecolor{saffron}{rgb}{0.95,0.75,0.2}
\definecolor{turquoise}{rgb}{0.0,0.5,0.5}
\definecolor{dark_green}{rgb}{0, 0.5, 0}
\definecolor{purple}{rgb}{0.65,0,0.65}
\newcommand{\rz}[1]{{\color{black}#1}}

\usepackage[pagebackref=true,breaklinks=true,letterpaper=true,colorlinks,bookmarks=false]{hyperref}

\iccvfinalcopy 

\def\iccvPaperID{1281} 

\ificcvfinal\fi
\begin{document}

\title{DualGAN: Unsupervised Dual Learning for Image-to-Image Translation}

%

\author[1,2]{Zili Yi}
\author[2]{Hao Zhang}
\author[2]{Ping Tan}
\author[1]{Minglun Gong}
\affil[1]{Memorial University of Newfoundland, Canada}
\affil[2]{Simon Fraser University, Canada}

%
%
%


\maketitle
\begin{abstract}

Conditional Generative Adversarial Networks (GANs) for cross-domain image-to-image translation have made much progress recently~\cite{ledig2016photo,li2016precomputed,wang2016generative,mathieu2015deep,isola2016image,taigman2016unsupervised}. Depending 
on the task complexity, thousands to millions of labeled image pairs are needed to train a conditional GAN. However, 
human labeling is expensive, even impractical, and large quantities of data may not always be available. Inspired by dual learning from 
natural language translation~\cite{xia2016dual}, we develop a novel {\em dual-GAN\/} mechanism, which enables image translators to be trained 
from two sets of {\em unlabeled\/} images from two domains. In our architecture, the primal GAN learns to translate images from 
domain $U$ to those in domain $V$, while the dual GAN learns to invert the task. The closed loop made by the primal and dual tasks allows 
images from either domain to be translated and then reconstructed. Hence a loss function that accounts for the reconstruction error of images 
can be used to train the translators. Experiments on multiple image translation tasks with unlabeled data show considerable performance gain 
of DualGAN over a single GAN. For some tasks, DualGAN can even achieve comparable or slightly better results than conditional GAN trained 
on fully labeled data.
\end{abstract}

\section{Introduction}
\label{sec:intro}

Many image processing and computer vision tasks, e.g., image segmentation, 
stylization, and abstraction, can be posed as image-to-image translation problems~\cite{isola2016image}, 
which convert one visual representation of an object or scene into another. Conventionally, these tasks have 
been tackled separately due to their intrinsic 
disparities~\cite{ledig2016photo,li2016precomputed,wang2016generative,mathieu2015deep,isola2016image,taigman2016unsupervised}. 
It is not until the past two years that general-purpose and end-to-end deep learning frameworks, most notably 
those utilizing fully convolutional networks (FCNs)~\cite{long2015fully} and conditional generative adversarial 
nets (cGANs)~\cite{isola2016image}, have been developed to enable a {\em unified\/} treatment of these tasks.

Up to date, these general-purpose methods have all been supervised and trained with a large number of
 {\em labeled\/} and {\em matching\/} image {\em pairs\/}. In practice however, acquiring
such training data can be time-consuming (e.g., with pixelwise or patchwise labeling) and even unrealistic.
For example, while there are plenty of photos or sketches available, photo-sketch image pairs depicting the 
same people under the same pose are scarce. In other image translation settings, e.g., converting daylight 
scenes to night scenes, even though labeled and matching image pairs can be obtained with stationary 
cameras, moving objects in the scene often cause 
varying degrees of content discrepancies.

In this paper, we aim to develop an {\em unsupervised\/} learning framework for general-purpose
image-to-image translation, which only relies on {\em unlabeled\/} image data, such as two sets of 
photos and sketches for the photo-to-sketch conversion task. The obvious technical challenge is
how to train a translator without any data characterizing correct translations. Our approach is inspired
by {\em dual learning\/} from natural language processing~\cite{xia2016dual}. Dual learning trains 
{\em two\/} ``opposite'' language translators (e.g., English-to-French and French-to-English) 
simultaneously by minimizing the {\em reconstruction loss\/} resulting from a {\em nested\/} application 
of the two translators. The two translators represent a primal-dual pair and the nested 
application forms a closed loop, allowing the application of reinforcement learning. Specifically, the
reconstruction loss measured over monolingual data (either English or French) would generate informative 
feedback to train a bilingual translation model.



Our work develops a dual learning framework for image-to-image translation for the first time and differs
from the original NLP dual learning method of Xia et al.~\cite{xia2016dual} in two main aspects. First, 
the NLP method relied on pre-trained (English and French) language models to indicate how confident 
the the translator outputs are natural sentences in their respective target languages. With 
general-purpose processing in mind and the realization that such pre-trained models are difficult to 
obtain for many image translation tasks, our work develops GAN discriminators~\cite{goodfellow2014generative} 
that are trained adversarially with the translators to capture domain distributions. Hence, we call our
learning architecture {\em DualGAN\/}. Furthermore, we employ FCNs as translators which naturally accommodate the 2D structure of images, rather than sequence-to-sequence translation models such as LSTM or Gated Recurrent Unit (GUT).

Taking two sets of unlabeled images as input, each characterizing an image domain, DualGAN simultaneously 
learns two reliable image translators from one domain to the other and hence can operate on a wide variety of 
image-to-image translation tasks. The effectiveness of DuanGAN is validated through comparison with both 
GAN (with an image-conditional generator and the original discriminator) and conditional GAN~\cite{isola2016image}. 
The comparison results demonstrate that, for some applications, DualGAN can outperform supervised methods trained on labeled data.

\section{Related work}


Since the seminal work by Goodfellow et al.~\cite{goodfellow2014generative} in 2014, a series of
GAN-family methods have been proposed for a wide variety of problems. The original GAN can learn 
a generator to capture the distribution of real data by introducing an adversarial discriminator that 
evolves to discriminate between the real data and the fake~\cite{goodfellow2014generative}. Soon
after, various conditional GANs (cGAN) have been proposed to condition the image generation on 
class labels~\cite{mirza2014conditional}, attributes~\cite{perarnau2016invertible,yan2016attribute2image}, 
texts~\cite{reed2016generative}, and 
images~\cite{ledig2016photo,li2016precomputed,wang2016generative,mathieu2015deep,isola2016image,taigman2016unsupervised}. 

Most image-conditional models were developed for specific applications such as 
super-resolution~\cite{ledig2016photo}, texture synthesis~\cite{li2016precomputed}, style transfer from 
normal maps to images~\cite{wang2016generative}, and video prediction~\cite{mathieu2015deep}, 
whereas few others were aiming for general-purpose processing~\cite{isola2016image,taigman2016unsupervised}. 
The general-purpose solution for image-to-image translation proposed by Isola et al.~\cite{isola2016image} 
requires significant number of labeled image pairs. The unsupervised mechanism for cross-domain image 
conversion presented by Taigman et al.~\cite{taigman2016unsupervised} can train an image-conditional 
generator without paired images, but relies on a sophisticated pre-trained function that maps images from either 
domain to an intermediate representation, which requires labeled data in other formats. 



\textbf{Dual learning} was first proposed by Xia et al.~\cite{xia2016dual} to reduce the requirement on labeled 
data in training English-to-French and French-to-English translators. The French-to-English translation is the dual 
task to English-to-French translation, and they can be trained side-by-side. The key idea of dual learning is to set 
up a dual-learning game which involves two agents, each of whom only understands one language, and can evaluate 
how likely the translated are natural sentences in targeted language and to what extent the reconstructed are consistent 
with the original. Such a mechanism is played alternatively on both sides, allowing translators to be trained from monolingual data only. 

Despite of a lack of parallel bilingual data, two types of feedback signals can be generated: the 
membership score which evaluates the likelihood of the translated texts belonging to the targeted language, and the 
reconstruction error that measures the disparity between the reconstructed sentences and the original. Both signals 
are assessed with the assistance of application-specific domain knowledge, i.e.,
the pre-trained English and French language models. 

In our work, we aim for a general-purpose solution for image-to-image conversion and hence do not utilize 
any domain-specific knowledge or pre-trained domain representations. Instead, we use a domain-adaptive
GAN discriminator to evaluate the membership score of translated samples, whereas the reconstruction error 
is measured as the mean of absolute difference between the reconstructed and original images within 
each image domain.

\rz{
In CycleGAN, a concurrent work by Zhu et al.~\cite{zhu2017CycleGAN}, the same idea for unpaired
image-to-image translation is proposed, where the primal-dual relation in DualGAN is referred to as a
cyclic mapping and their {\em cycle consistency loss\/} is essentially the same as our reconstruction loss.
Superiority of CycleGAN has been demonstrated on several tasks where paired training data hardly exist, 
e.g., in object transfiguration and painting style and season transfer.

Recent work by Liu and Tuzel~\cite{liu2016coupled}, which we refer to as coupled GAN or CoGAN, also 
trains two GANs together to solve image translation problems without paired training data. Unlike DualGAN
or CycleGAN, the two GANs in CoGAN are not linked to enforce cycle consistency. Instead, CoGAN learns 
a joint distribution over images from two domains. 
By sharing weight parameters corresponding to high level semantics in both generative and 
discriminative networks, CoGAN can enforce the two GANs to interpret these image semantics in the same 
way. However, the weight-sharing assumption in CoGAN and similar approaches, e.g.,~\cite{aytar2016,liu2017},
does not lead to effective general-purpose solutions as its applicability is task-dependent, leading to unnatural
image translation results, as shown in comparative studies by CycleGAN~\cite{zhu2017CycleGAN}.}

DualGAN and CycleGAN both aim for general-purpose image-to-image translations without requiring
a joint representation to bridge the two image domains. In addition, DualGAN trains both primal and dual 
GANs at the same time, allowing a reconstruction error term to be used to generate informative 
feedback signals. 

\section{Method}

\begin{figure*}
\begin{center}
\includegraphics[width=0.98\linewidth]{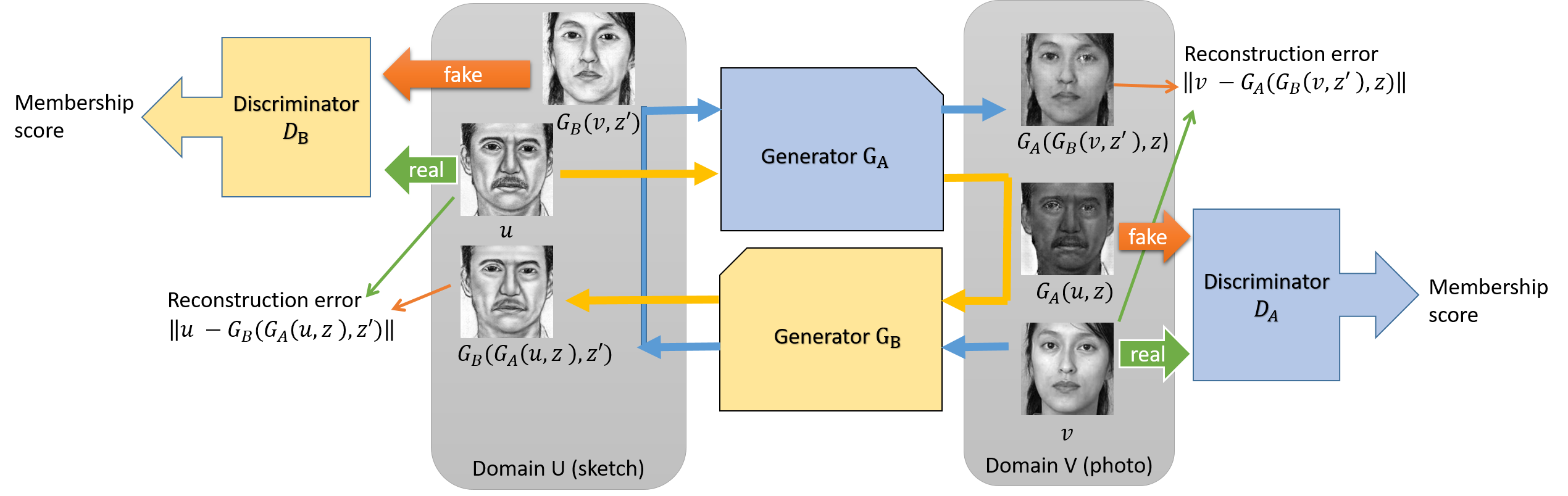}
\caption{Network architecture and data flow chart of DualGAN for image-to-image translation.}
\label{fig:framework}
\end{center}
\end{figure*}

Given two sets of unlabeled and unpaired images sampled from domains $U$ and $V$, respectively, the primal task
of DualGAN is to learn a generator $G_A: U \rightarrow V$ that maps an image $u \in U$ to an image $v \in V$,
while the dual task is to train an inverse generator $G_B: V \rightarrow U$. To realize this, we employ two GANs, 
the primal GAN and the dual GAN. The primal GAN learns the generator $G_A$ and a discriminator $D_A$ that 
discriminates between $G_A$'s fake outputs and real members of domain $V$. Analogously, the dual GAN learns 
the generator $G_B$ and a discriminator $D_B$. The overall architecture and data flow are illustrated in Fig.~\ref{fig:framework}. 

As shown in Fig.~\ref{fig:framework}, image $u \in U$ is translated to domain $V$ using $G_A$. How well the 
translation $G_A(u,z)$ fits in $V$ is evaluated by $D_A$, where $z$ is random noise, and so is $z'$ that appears 
below. $G_A(u,z)$ is then translated back to domain $U$ using $G_B$, which outputs $G_B(G_A(u,z),z')$ as the 
reconstructed version of $u$. Similarly, $v \in V$ is translated to $U$ as $G_B(v,z')$ and then reconstructed as 
$G_A(G_B(v,z'),z)$. The discriminator $D_A$ is trained with $v$ as positive samples and $G_A(u,z)$ as negative examples, 
whereas $D_B$ takes $u$ as positive and $G_B(v,z')$ as negative. Generators $G_A$ and $G_B$ are optimized to emulate 
``fake'' outputs to blind the corresponding discriminators $D_A$ and $D_B$, as well as to minimize the two {\em reconstruction losses\/}
$\|G_A(G_B(v,z'),z) - v\|$ and $\|G_B(G_A(u,z),z') - u\|$. 


\subsection{Objective}

As in the traditional GAN, the objective of discriminators is to discriminate the generated fake samples from the real ones. 
Nevertheless, here we use the loss format advocated by Wasserstein GAN (WGAN)~\cite{arjovsky2017wasserstein} rather 
than the sigmoid cross-entropy loss used in the original GAN~\cite{goodfellow2014generative}. It is proven that the former 
performs better in terms of generator convergence and sample quality, as well as in improving the stability of the 
optimization~\cite{arjovsky2017wasserstein}. The corresponding loss functions used in $D_A$ and $D_B$ are defined as:

\begin{eqnarray}
l^d_A(u,v) = D_A(G_A (u,z)) - D_A(v), \label{eq:loss_da} \\
l^d_B(u,v) = D_B(G_B (v,z')) - D_B(u),  \label{eq:loss_db}
\end{eqnarray}
where $u \in U$ and $v \in V$.

The same loss function is used for both generators $G_A$ and $G_B$ as they share the same objective. Previous works on 
conditional image synthesis found it beneficial to replace $L_2$ distance with $L_1$, since the former often leads 
to blurriness~\cite{larsen2015autoencoding,xia2016dual}. Hence, we adopt $L_1$ distance to measure the recovery error, 
which is added to the GAN objective to force the translated samples to obey the domain distribution: 
\begin{equation}
\label{eq:loss_g}
\begin{gathered}
l^g(u,v)=\lambda_U\|u-G_B(G_A(u,z),z')\| + \\
\lambda_V\|v-G_A(G_B(v,z'),z)\| \\
  -D_B(G_{B}(v,z'))-D_A(G_{A}(u,z)), 
\end{gathered}
\end{equation}
where $u \in U$, $v \in V$, and $\lambda_U$, $\lambda_V$ are two constant parameters. Depending on the application, 
$\lambda_U$ and $\lambda_V$ are typically set to a value within $[100.0, 1,000.0]$. If $U$ contains natural images and 
$V$ does not (e.g., aerial photo-maps), we find it more effective to use smaller $\lambda_U$ than $\lambda_V$.
%

\subsection{Network configuration}

DualGAN is constructed with identical network architecture for $G_A$ and $G_B$. The generator is configured with equal number of 
downsampling (pooling) and upsampling layers. In addition, we configure the generator with skip connections 
between mirrored downsampling and upsampling layers as in~\cite{ronneberger2015u, isola2016image}, making it 
a U-shaped net. 
Such a design enables low-level information to be shared between input and output, which is beneficial since many image 
translation problems implicitly assume alignment between image structures in the input and output (e.g., object shapes, textures, 
clutter, etc.). Without the skip layers, information from all levels has to pass through the bottleneck, typically causing significant 
loss of high-frequency information. 
Furthermore, similar to \cite{isola2016image}, we did not explicitly provide the noise vectors $z$, $z'$. Instead, they are provided only 
in the form of dropout and applied to several layers of our generators at both training and test phases.

For discriminators, we employ the Markovian PatchGAN architecture as explored in~\cite{li2016precomputed}, which assumes 
independence between pixels distanced beyond a specific patch size and models images only at the patch level rather than over
the full image. Such a configuration is effective in capturing local high-frequency features such as texture and style, but less so 
in modeling global distributions. It fulfills our needs well, since the recovery loss encourages preservation of global and low-frequency 
information and the discriminators are designated to capture local high-frequency information. The effectiveness of this 
configuration has been verified on various translation tasks~\cite{xia2016dual}. Similar to~\cite{xia2016dual}, we run this discriminator 
convolutionally across the image, averaging all responses to provide the ultimate output. An extra advantage of such a scheme is 
that it requires fewer parameters, runs faster, and has no constraints over the size of the input image. The patch size at which the 
discriminator operates is fixed at $70\times 70$, and the image resolutions were mostly $256 \times 256$, same as pix2pix~\cite{isola2016image}.

\subsection{Training procedure}

To optimize the DualGAN networks, we follow the training procedure proposed in WGAN~\cite{arjovsky2017wasserstein}; see 
Alg.~\ref{alg:procedure}. We train the discriminators $n_{critic}$ steps, then one step on generators. We employ mini-batch Stochastic 
Gradient Descent and apply the RMSProp solver, as momentum based methods such as Adam would occasionally cause 
instability~\cite{arjovsky2017wasserstein}, and RMSProp is known to perform well even on highly non-stationary problems~\cite{tieleman2012lecture, arjovsky2017wasserstein}. We typically set the number of critic iterations per generator iteration $n_{critic}$ to $2$-$4$ and assign batch size to $1$-$4$, without
noticeable differences on effectiveness in the experiments. The clipping parameter $c$ is normally set in $[0.01, 0.1]$, 
varying by application.

\begin{algorithm}
\caption{DualGAN training procedure}
\label{alg:procedure}
\begin{algorithmic}[1]
\REQUIRE Image set $U$, image set $V$, GAN $A$ with generator parameters $\theta_A$ and discriminator parameters $\omega_A$, GAN $B$ with generator parameters $\theta_B$ and discriminator parameters $\omega_B$, clipping parameter $c$, batch size $m$, and $n_{critic}$
	\STATE Randomly initialize $\omega_i$, $\theta_i$, $i \in \{A,B\}$
	\REPEAT 
		\FOR {$t=1,\ldots,n_{critic}$}
			\STATE sample images $\{u^{(k)}\}_{k=1}^{m} \subseteq U$, $\{v^{(k)}\}_{k=1}^{m} \subseteq V$
			\STATE update $\omega_A$ to minimize $\frac{1}{m}\sum_{k=1}^m l^d_A(u^{(k)},v^{(k)})$
			\STATE update $\omega_B$ to minimize $\frac{1}{m}\sum_{k=1}^m l^d_B(u^{(k)},v^{(k)})$
			\STATE $clip(\omega_A,-c,c)$, $clip(\omega_B,-c,c)$
		\ENDFOR
		\STATE sample images $\{u^{(k)}\}_{k=1}^{m} \subseteq U$, $\{v^{(k)}\}_{k=1}^{m} \subseteq V$
		\STATE update $\theta_A$, $\theta_B$ to minimize  $\frac{1}{m}\sum_{k=1}^m l^g(u^{(k)},v^{(k)})$
	\UNTIL{convergence}
\end{algorithmic}
\end{algorithm}

Training for traditional GANs needs to carefully balance between the generator and the discriminator, since,
as the discriminator improves, the sigmoid cross-entropy loss is locally saturated and may lead to vanishing 
gradients. Unlike in traditional GANs, the Wasserstein loss is differentiable almost everywhere, resulting in a better discriminator. 
At each iteration, the generators are not trained until the discriminators have been trained for $n_{critic}$ steps. Such a procedure 
enables the discriminators to provide more reliable gradient information~\cite{arjovsky2017wasserstein}.

\section{Experimental results and evaluation}
\label{sect:exp}

To assess the capability of DualGAN in general-purpose image-to-image translation, we 
conduct experiments on a variety of tasks, including photo-sketch conversion, 
label-image translation, and artistic stylization.

\begin{figure}
\begin{center}
\includegraphics[width=0.19\linewidth]{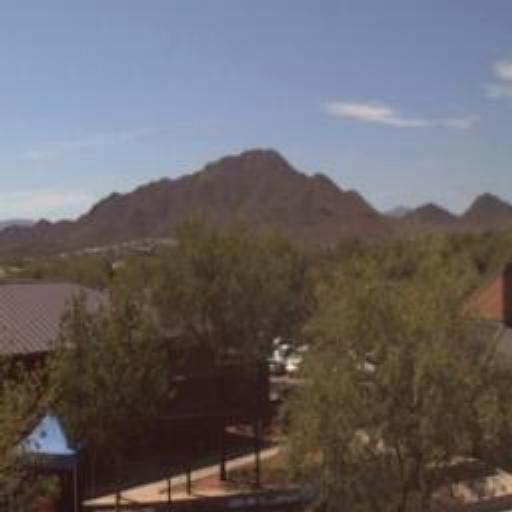}
\includegraphics[width=0.19\linewidth]{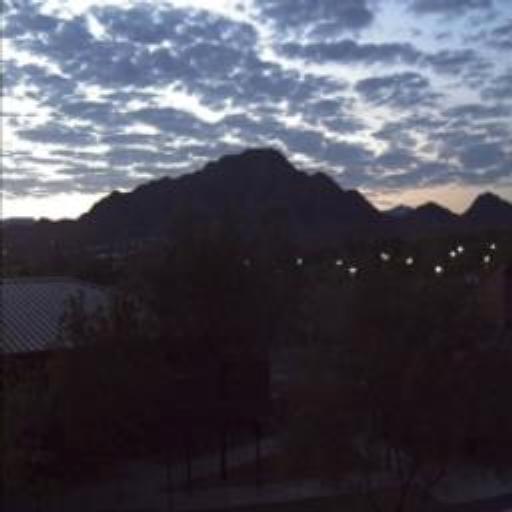}
\includegraphics[width=0.19\linewidth]{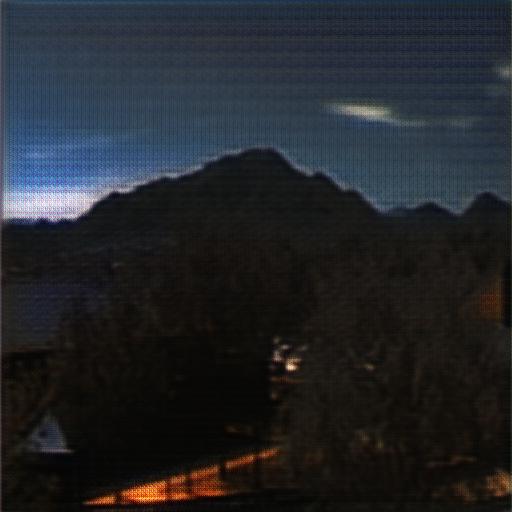}
\includegraphics[width=0.19\linewidth]{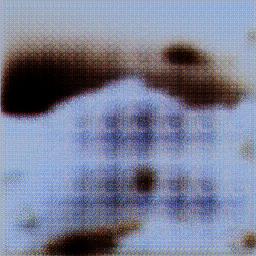}
\includegraphics[width=0.19\linewidth]{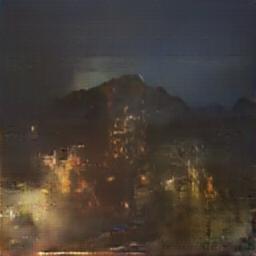}

\includegraphics[width=0.19\linewidth]{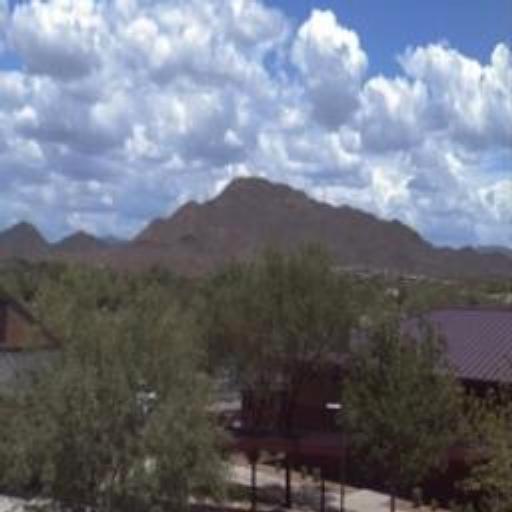}
\includegraphics[width=0.19\linewidth]{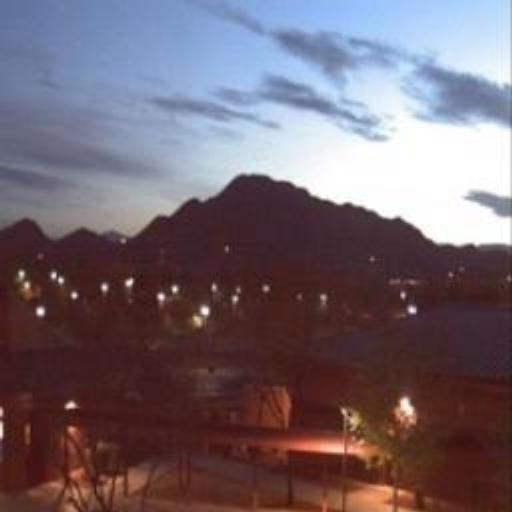}
\includegraphics[width=0.19\linewidth]{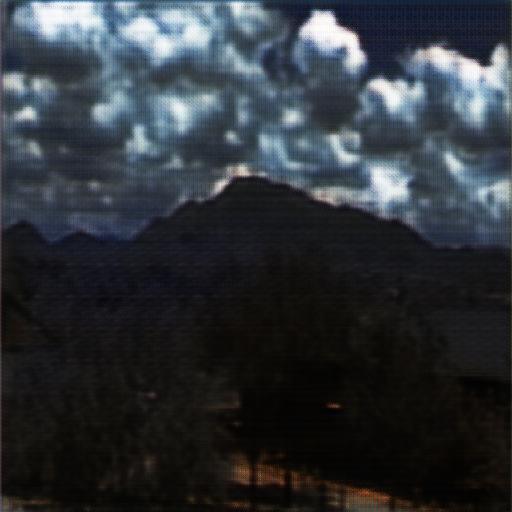}
\includegraphics[width=0.19\linewidth]{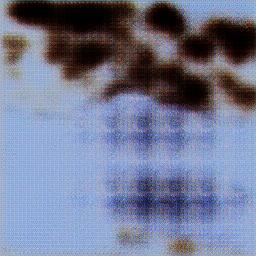}
\includegraphics[width=0.19\linewidth]{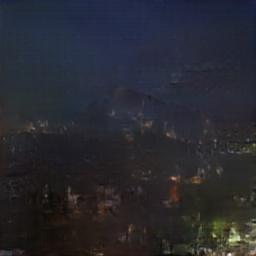}

\includegraphics[width=0.19\linewidth]{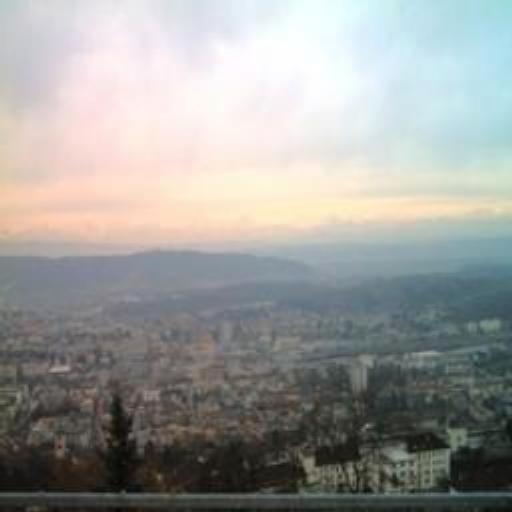}
\includegraphics[width=0.19\linewidth]{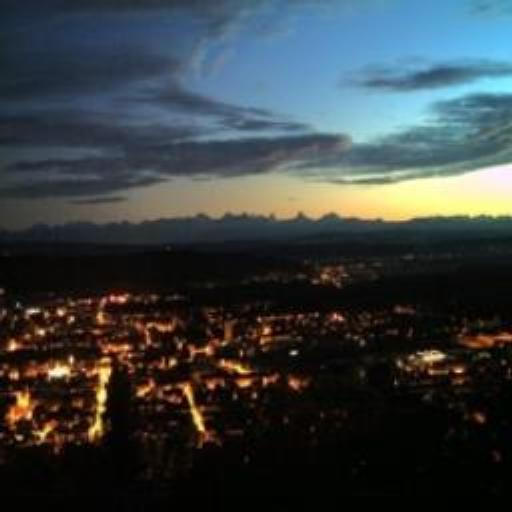}
\includegraphics[width=0.19\linewidth]{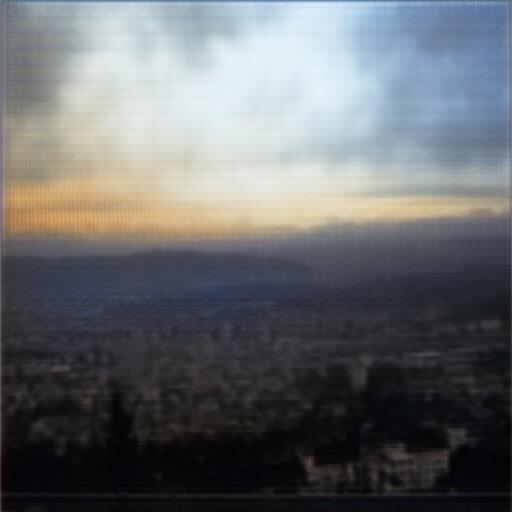}
\includegraphics[width=0.19\linewidth]{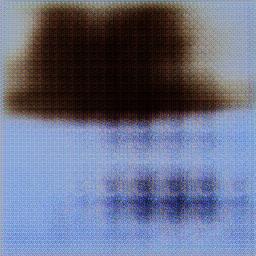}
\includegraphics[width=0.19\linewidth]{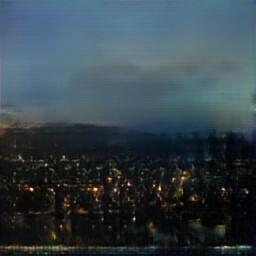}

\includegraphics[width=0.19\linewidth]{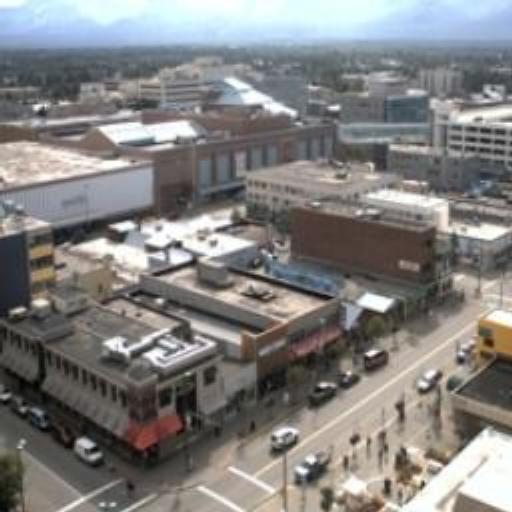}
\includegraphics[width=0.19\linewidth]{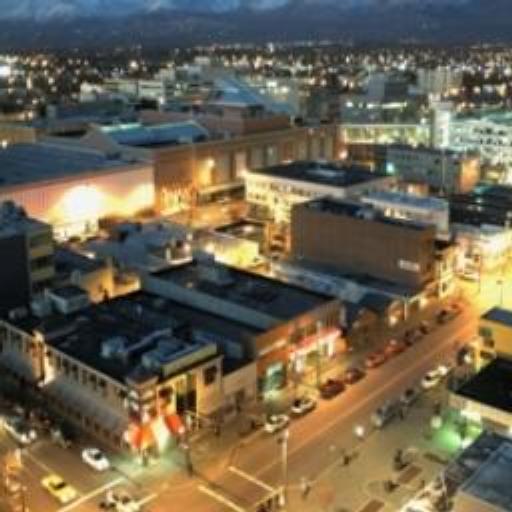}
\includegraphics[width=0.19\linewidth]{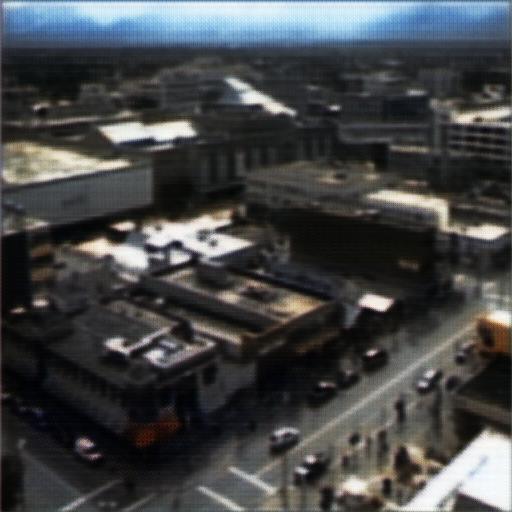}
\includegraphics[width=0.19\linewidth]{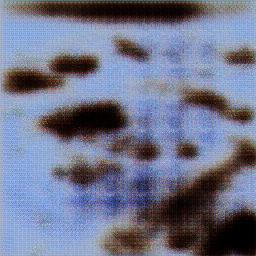}
\includegraphics[width=0.19\linewidth]{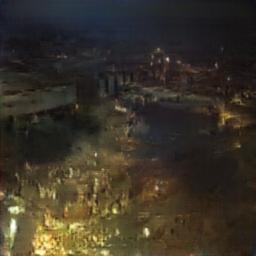}

\begin{subfigure}[]{0.19\linewidth}\caption*{Input}\end{subfigure}
\begin{subfigure}[]{0.19\linewidth}\caption*{GT}\end{subfigure}
\begin{subfigure}[]{0.19\linewidth}\caption*{\textbf{DualGAN}}\end{subfigure}
\begin{subfigure}[]{0.19\linewidth}\caption*{GAN}\end{subfigure}
\begin{subfigure}[]{0.19\linewidth}\caption*{cGAN~\cite{isola2016image}}\end{subfigure}
\caption{Results of day$\rightarrow$night translation. cGAN~\cite{isola2016image} is trained with labeled data, whereas DualGAN and GAN are trained in an unsupervised manner. DualGAN successfully emulates the night scenes while preserving textures in the inputs, e.g., see differences over the cloud regions between our results and the ground truth (GT). In comparison, results of cGAN and GAN contain much less details.}
\label{fig:day}
\end{center}
\end{figure}

\begin{figure}
\begin{center}
\includegraphics[width=0.19\linewidth]{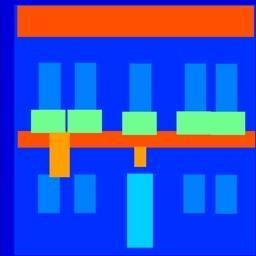}
\includegraphics[width=0.19\linewidth]{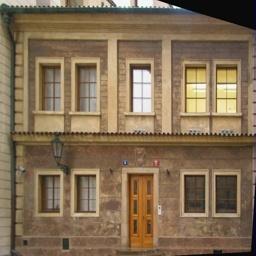}
\includegraphics[width=0.19\linewidth]{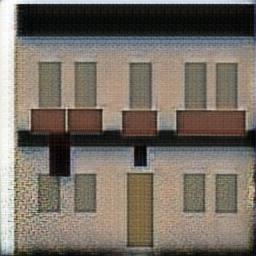}
\includegraphics[width=0.19\linewidth]{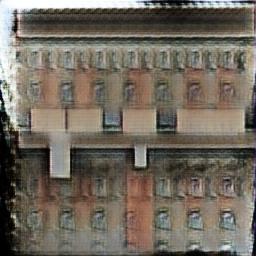}
\includegraphics[width=0.19\linewidth]{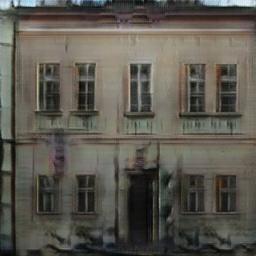}

\includegraphics[width=0.19\linewidth]{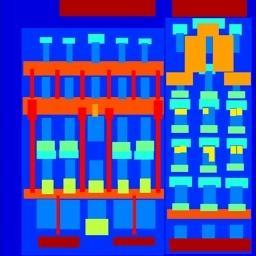}
\includegraphics[width=0.19\linewidth]{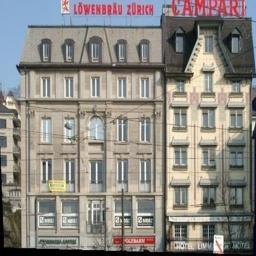}
\includegraphics[width=0.19\linewidth]{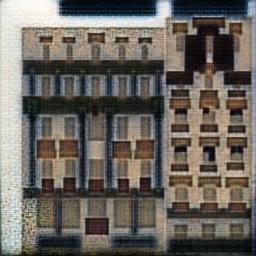}
\includegraphics[width=0.19\linewidth]{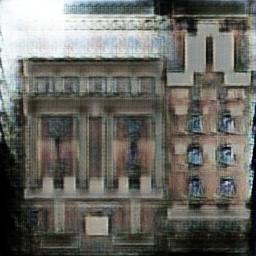}
\includegraphics[width=0.19\linewidth]{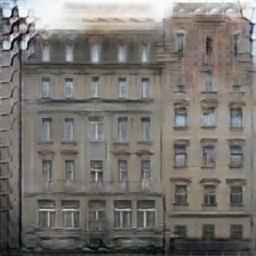}

\includegraphics[width=0.19\linewidth]{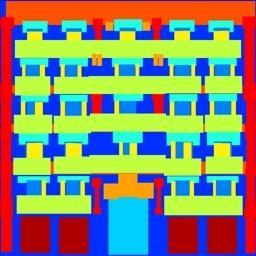}
\includegraphics[width=0.19\linewidth]{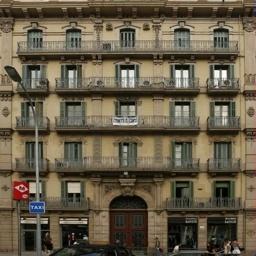}
\includegraphics[width=0.19\linewidth]{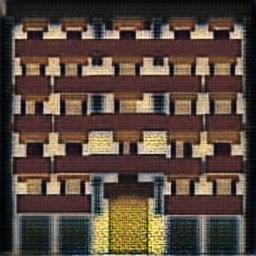}
\includegraphics[width=0.19\linewidth]{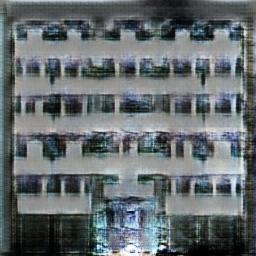}
\includegraphics[width=0.19\linewidth]{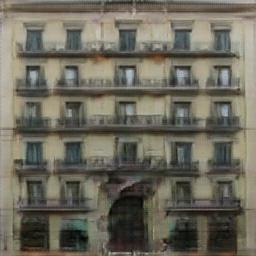}

\includegraphics[width=0.19\linewidth]{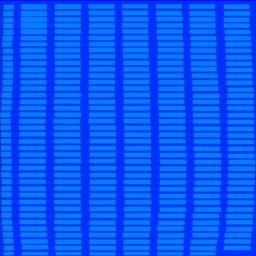}
\includegraphics[width=0.19\linewidth]{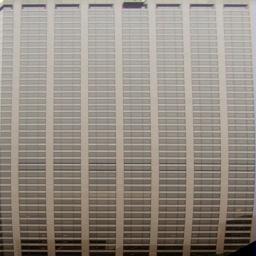}
\includegraphics[width=0.19\linewidth]{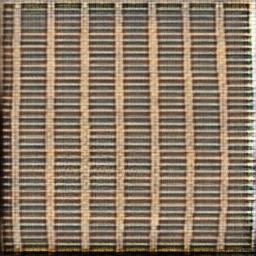}
\includegraphics[width=0.19\linewidth]{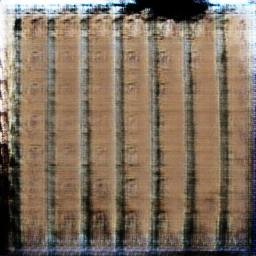}
\includegraphics[width=0.19\linewidth]{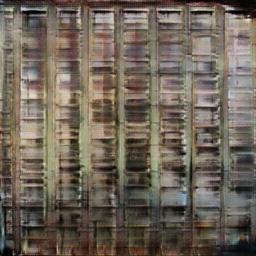}

\begin{subfigure}[]{0.19\linewidth}\caption*{Input}\end{subfigure}
\begin{subfigure}[]{0.19\linewidth}\caption*{GT}\end{subfigure}
\begin{subfigure}[]{0.19\linewidth}\caption*{\textbf{DualGAN}}\end{subfigure}
\begin{subfigure}[]{0.19\linewidth}\caption*{GAN}\end{subfigure}
\begin{subfigure}[]{0.19\linewidth}\caption*{cGAN~\cite{isola2016image}}\end{subfigure}
\caption{Results of label$\rightarrow$facade translation. DualGAN faithfully preserves the structures 
in the label images, even though some labels do not match well with the corresponding photos
in finer details. In contrast, results from GAN and cGAN contain many artifacts. Over regions with
label-photo misalignment, cGAN often yields blurry output (e.g., the roof in second row and 
the entrance in third row).}
\label{fig:facades}
\end{center}
\end{figure}

To compare DualGAN with GAN and cGAN~\cite{isola2016image}, four labeled datasets are used: 
PHOTO-SKETCH~\cite{wang2009face,zhang2011coupled}, DAY-NIGHT~\cite{laffont2014transient}, 
LABEL-FACADES~\cite{tylevcek2013spatial}, and AERIAL-MAPS, which was directly captured from 
Google Map~\cite{isola2016image}. These datasets consist of corresponding images between two 
domains; they serve as ground truth (GT) and can also be used for supervised learning. 
However, none of these datasets could guarantee accurate feature alignment at the pixel level. For example, 
the sketches in SKETCH-PHOTO dataset were drawn by artists and do not accurately align with the 
corresponding photos, moving objects and cloud pattern changes often show up in the DAY-NIGHT dataset, 
and the labels in LABEL-FACADES dataset are not always precise. This highlights, in part, the difficulty in 
obtaining high quality matching image pairs.

DualGAN enables us to utilize abundant unlabeled image sources from the Web. Two unlabeled and 
unpaired datasets are also tested in our experiments. The MATERIAL dataset includes images of objects made of 
different materials, e.g., stone, metal, plastic, fabric, and wood. These images were manually selected from 
Flickr and cover a variety of illumination conditions, compositions, color, texture, and material 
sub-types~\cite{sharan2009material}. This dataset was initially used for material recognition, but is applied 
here for material transfer. The OIL-CHINESE painting dataset includes artistic paintings of two disparate styles: 
oil and Chinese. All images were crawled from search engines and they contain images with varying 
quality, format, and size. We reformat, crop, and resize the images for training and evaluation.
In both of these datasets, no correspondence is available between images from different domains.

\section{Qualitative evaluation}

\begin{figure}
\begin{center}
\includegraphics[width=0.19\linewidth]{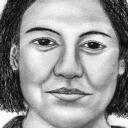}
\includegraphics[width=0.19\linewidth]{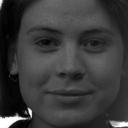}
\includegraphics[width=0.19\linewidth]{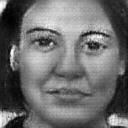}
\includegraphics[width=0.19\linewidth]{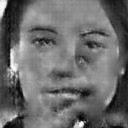}
\includegraphics[width=0.19\linewidth]{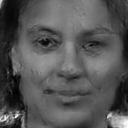}

\includegraphics[width=0.19\linewidth]{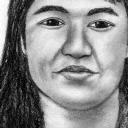}
\includegraphics[width=0.19\linewidth]{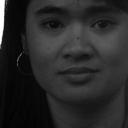}
\includegraphics[width=0.19\linewidth]{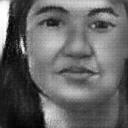}
\includegraphics[width=0.19\linewidth]{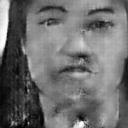}
\includegraphics[width=0.19\linewidth]{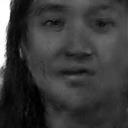}

\includegraphics[width=0.19\linewidth]{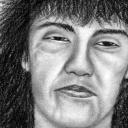}
\includegraphics[width=0.19\linewidth]{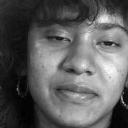}
\includegraphics[width=0.19\linewidth]{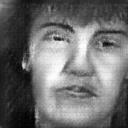}
\includegraphics[width=0.19\linewidth]{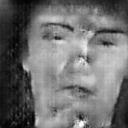}
\includegraphics[width=0.19\linewidth]{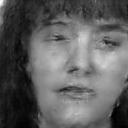}

\includegraphics[width=0.19\linewidth]{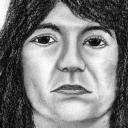}
\includegraphics[width=0.19\linewidth]{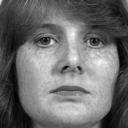}
\includegraphics[width=0.19\linewidth]{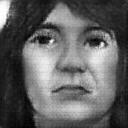}
\includegraphics[width=0.19\linewidth]{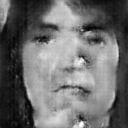}
\includegraphics[width=0.19\linewidth]{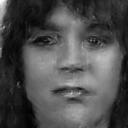}

\begin{subfigure}[]{0.19\linewidth}\caption*{Input}\end{subfigure}
\begin{subfigure}[]{0.19\linewidth}\caption*{GT}\end{subfigure}
\begin{subfigure}[]{0.19\linewidth}\caption*{\textbf{DualGAN}}\end{subfigure}
\begin{subfigure}[]{0.19\linewidth}\caption*{GAN}\end{subfigure}
\begin{subfigure}[]{0.19\linewidth}\caption*{cGAN~\cite{isola2016image}}\end{subfigure}
\caption{Photo$\rightarrow$sketch translation for faces. Results of DualGAN are 
generally sharper than those from cGAN, even though the former was trained using 
unpaired data, whereas the latter makes use of image correspondence.}
\label{fig:photo}
\end{center}
\end{figure}

\begin{figure}
\begin{center}
\includegraphics[width=0.19\linewidth]{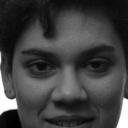}
\includegraphics[width=0.19\linewidth]{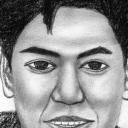}
\includegraphics[width=0.19\linewidth]{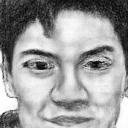}
\includegraphics[width=0.19\linewidth]{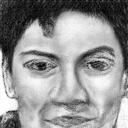}
\includegraphics[width=0.19\linewidth]{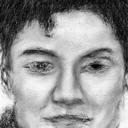}

\includegraphics[width=0.19\linewidth]{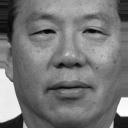}
\includegraphics[width=0.19\linewidth]{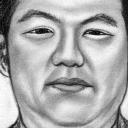}
\includegraphics[width=0.19\linewidth]{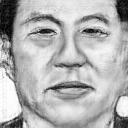}
\includegraphics[width=0.19\linewidth]{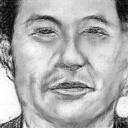}
\includegraphics[width=0.19\linewidth]{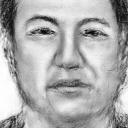}

\includegraphics[width=0.19\linewidth]{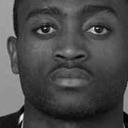}
\includegraphics[width=0.19\linewidth]{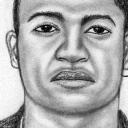}
\includegraphics[width=0.19\linewidth]{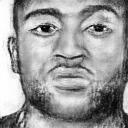}
\includegraphics[width=0.19\linewidth]{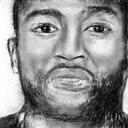}
\includegraphics[width=0.19\linewidth]{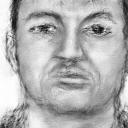}

\includegraphics[width=0.19\linewidth]{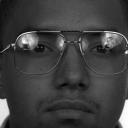}
\includegraphics[width=0.19\linewidth]{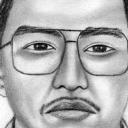}
\includegraphics[width=0.19\linewidth]{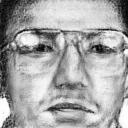}
\includegraphics[width=0.19\linewidth]{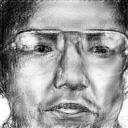}
\includegraphics[width=0.19\linewidth]{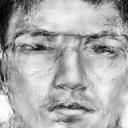}

\begin{subfigure}[]{0.19\linewidth}\caption*{Input}\end{subfigure}
\begin{subfigure}[]{0.19\linewidth}\caption*{GT}\end{subfigure}
\begin{subfigure}[]{0.19\linewidth}\caption*{\textbf{DualGAN}}\end{subfigure}
\begin{subfigure}[]{0.19\linewidth}\caption*{GAN}\end{subfigure}
\begin{subfigure}[]{0.19\linewidth}\caption*{cGAN~\cite{isola2016image}}\end{subfigure}
\caption{Results for sketch$\rightarrow$photo translation of faces. More 
artifacts and blurriness are showing up in results generated by GAN and cGAN than 
DualGAN.} \label{fig:sketch}
\end{center}
\end{figure}

\begin{figure}
\begin{center}
\includegraphics[width=0.19\linewidth]{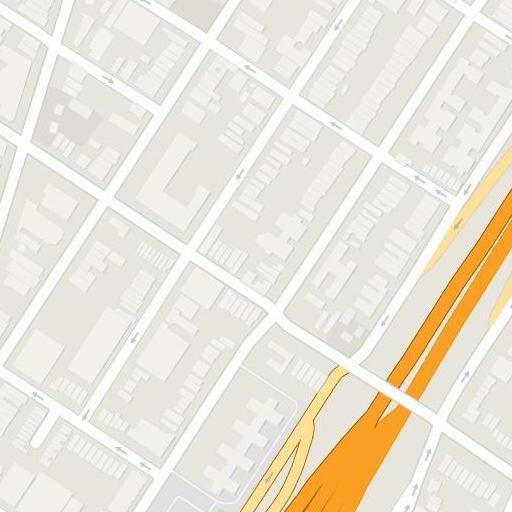}
\includegraphics[width=0.19\linewidth]{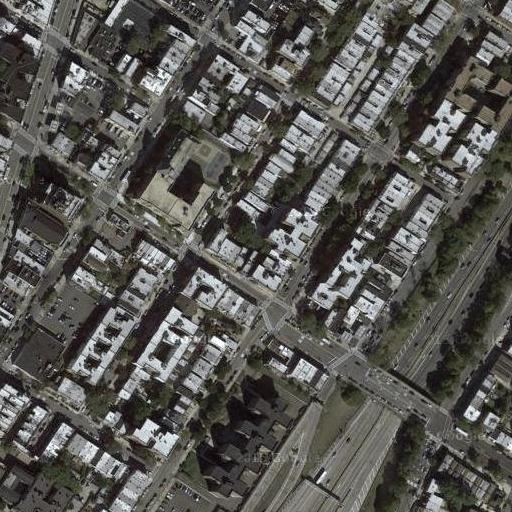}
\includegraphics[width=0.19\linewidth]{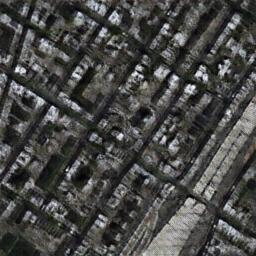}
\includegraphics[width=0.19\linewidth]{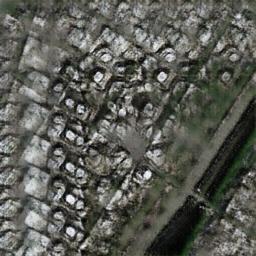}
\includegraphics[width=0.19\linewidth]{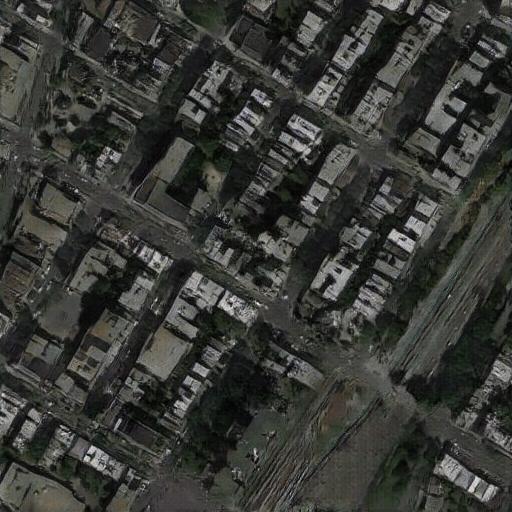}

\includegraphics[width=0.19\linewidth]{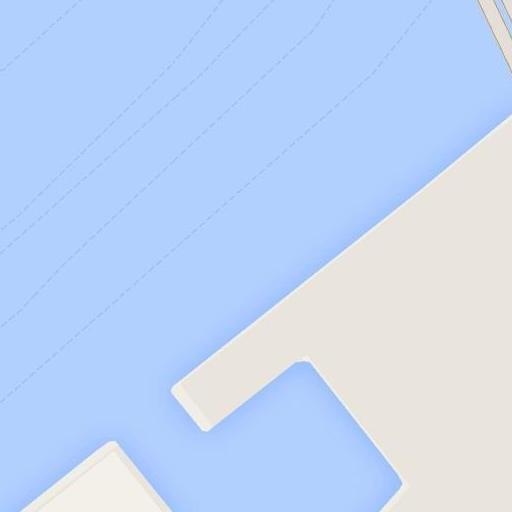}
\includegraphics[width=0.19\linewidth]{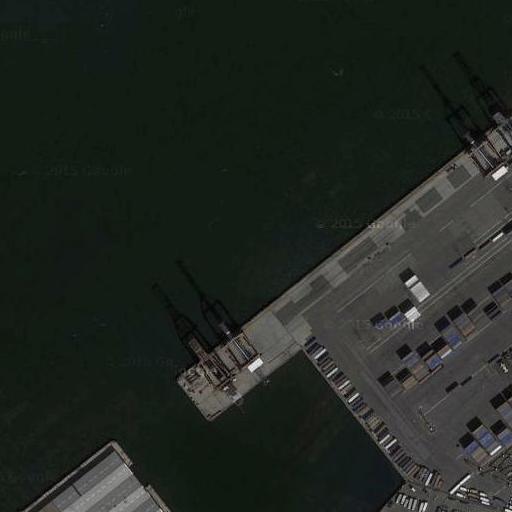}
\includegraphics[width=0.19\linewidth]{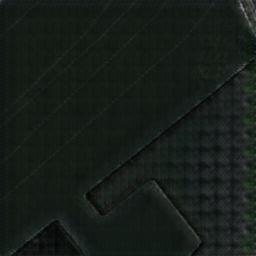}
\includegraphics[width=0.19\linewidth]{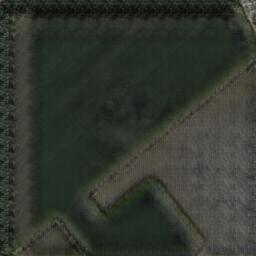}
\includegraphics[width=0.19\linewidth]{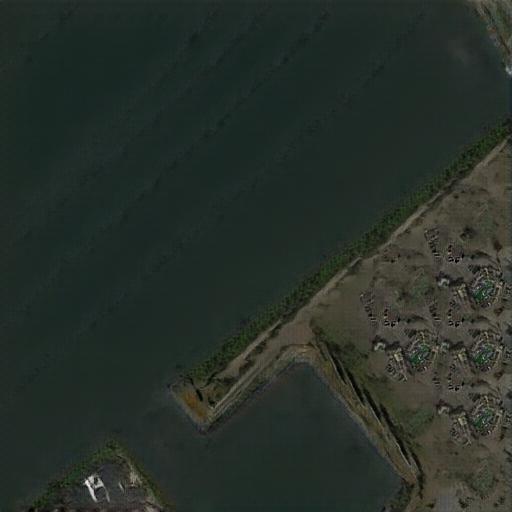}

\includegraphics[width=0.19\linewidth]{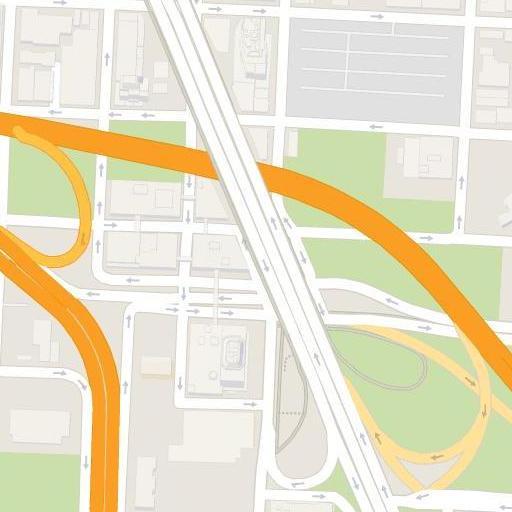}
\includegraphics[width=0.19\linewidth]{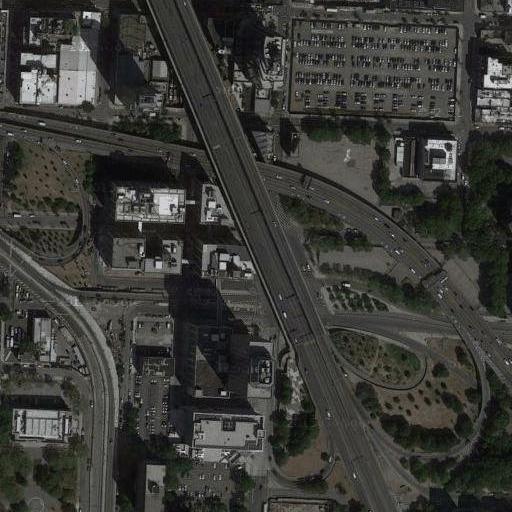}
\includegraphics[width=0.19\linewidth]{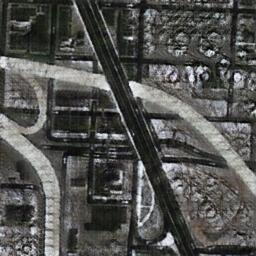}
\includegraphics[width=0.19\linewidth]{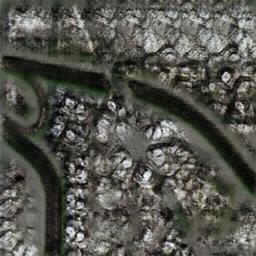}
\includegraphics[width=0.19\linewidth]{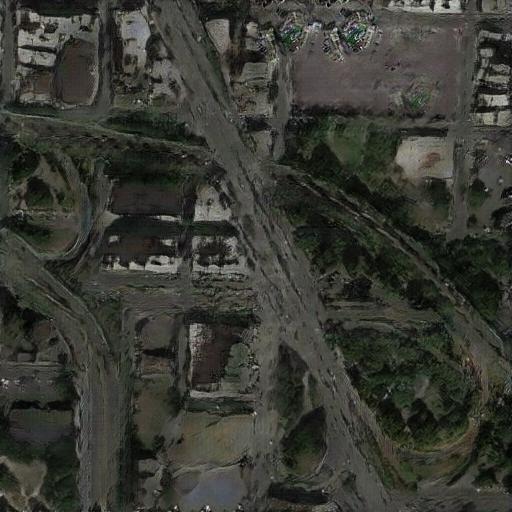}

\begin{subfigure}[]{0.19\linewidth}\caption*{Input}\end{subfigure}
\begin{subfigure}[]{0.19\linewidth}\caption*{GT}\end{subfigure}
\begin{subfigure}[]{0.19\linewidth}\caption*{\textbf{DualGAN}}\end{subfigure}
\begin{subfigure}[]{0.19\linewidth}\caption*{GAN}\end{subfigure}
\begin{subfigure}[]{0.19\linewidth}\caption*{cGAN~\cite{isola2016image}}\end{subfigure}
\caption{Map$\rightarrow$aerial photo translation. Without image correspondences for training, 
DualGAN may map the orange-colored interstate highways to building roofs with bright colors. 
Nevertheless, the DualGAN results are sharper than those from GAN and cGAN.}
\label{fig:maps}
\end{center}
\end{figure}

\begin{figure}
\begin{center}

\includegraphics[width=0.19\linewidth]{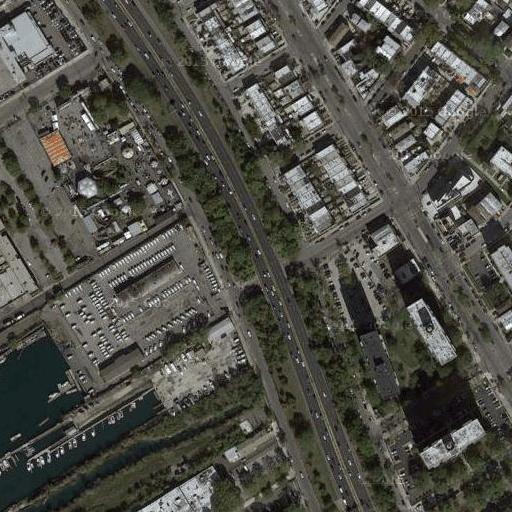}
\includegraphics[width=0.19\linewidth]{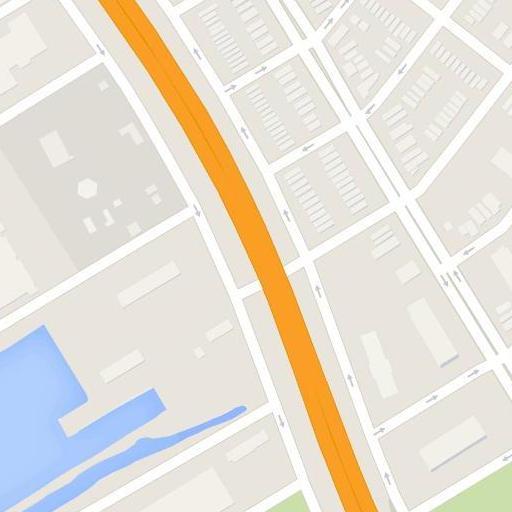}
\includegraphics[width=0.19\linewidth]{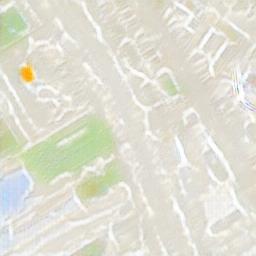}
\includegraphics[width=0.19\linewidth]{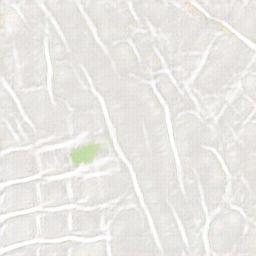}
\includegraphics[width=0.19\linewidth]{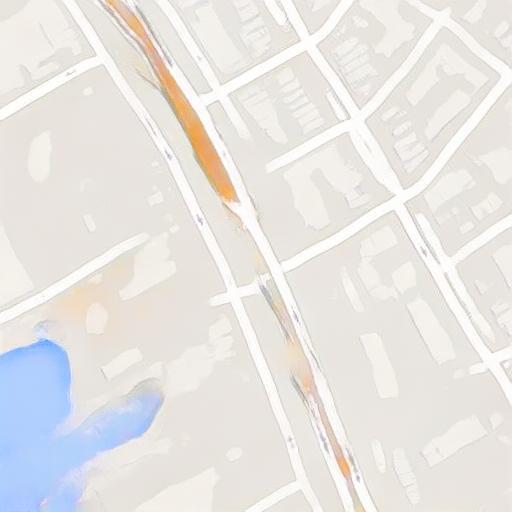}

\includegraphics[width=0.19\linewidth]{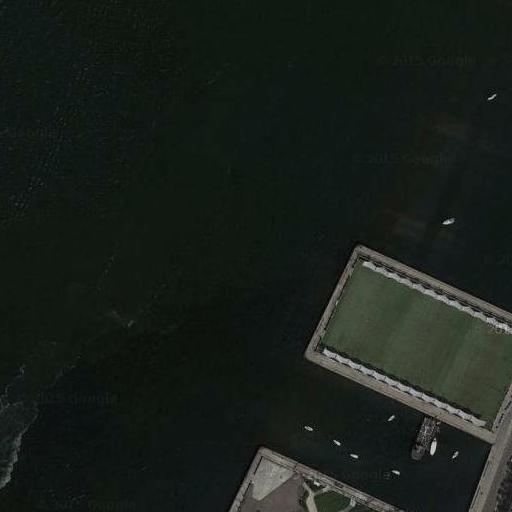}
\includegraphics[width=0.19\linewidth]{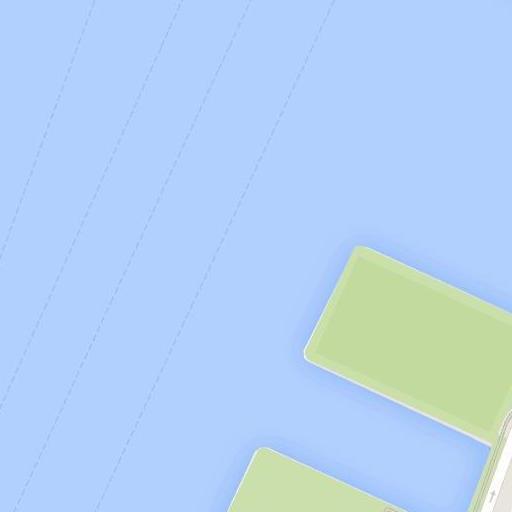}
\includegraphics[width=0.19\linewidth]{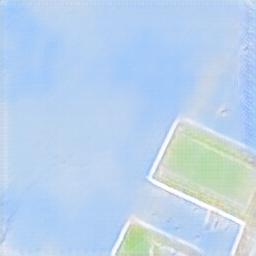}
\includegraphics[width=0.19\linewidth]{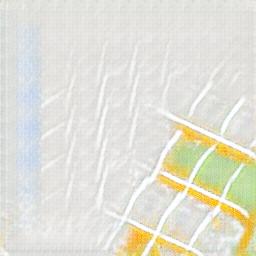}
\includegraphics[width=0.19\linewidth]{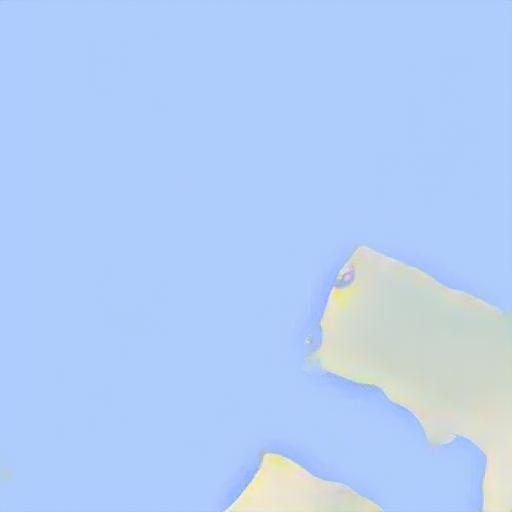}

\includegraphics[width=0.19\linewidth]{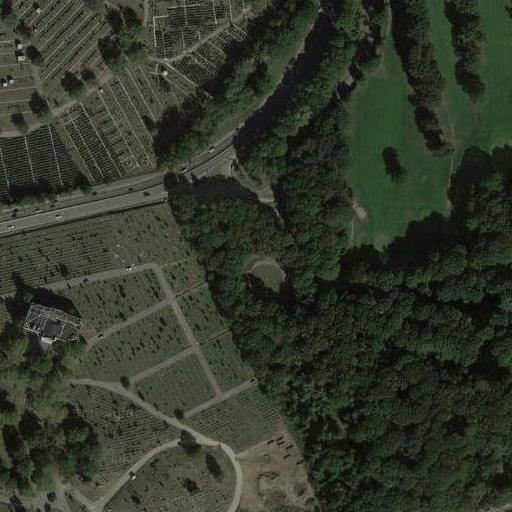}
\includegraphics[width=0.19\linewidth]{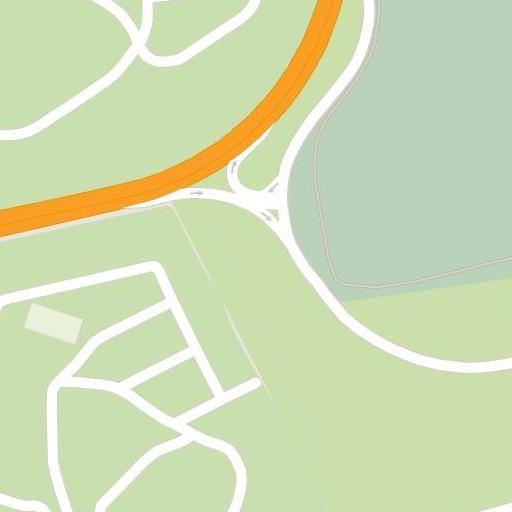}
\includegraphics[width=0.19\linewidth]{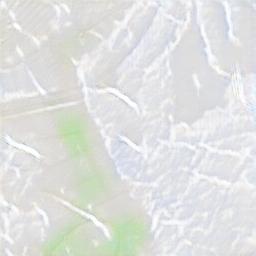}
\includegraphics[width=0.19\linewidth]{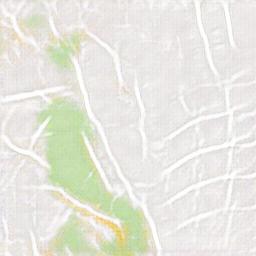}
\includegraphics[width=0.19\linewidth]{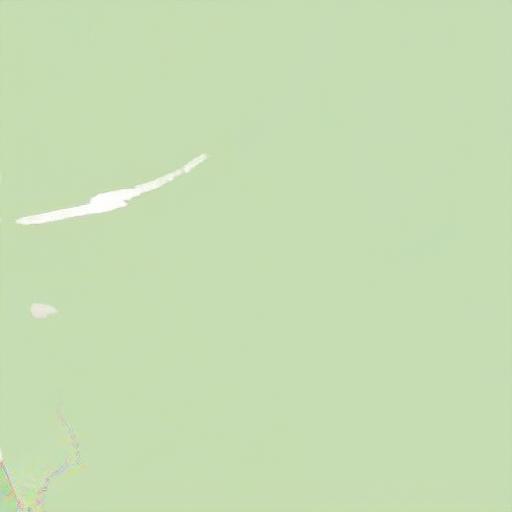}

\begin{subfigure}[]{0.19\linewidth}\caption*{Input}\end{subfigure}
\begin{subfigure}[]{0.19\linewidth}\caption*{GT}\end{subfigure}
\begin{subfigure}[]{0.19\linewidth}\caption*{\textbf{DualGAN}}\end{subfigure}
\begin{subfigure}[]{0.19\linewidth}\caption*{GAN)}\end{subfigure}
\begin{subfigure}[]{0.19\linewidth}\caption*{cGAN~\cite{isola2016image}}\end{subfigure}
\caption{Results for aerial photo$\rightarrow$map translation. DualGAN performs better than 
GAN, but not as good as cGAN. With additional pixel correspondence information, cGAN performs 
well in terms of labeling local roads, but still cannot detect interstate highways.}
\label{fig:aerial}
\end{center}
\end{figure}

\begin{figure}
\begin{center}
\includegraphics[width=0.19\linewidth]{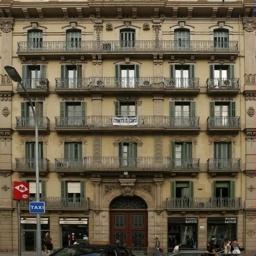}
\includegraphics[width=0.19\linewidth]{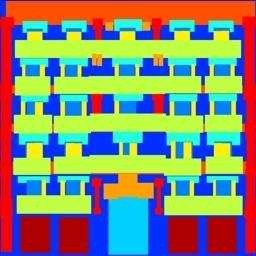}
\includegraphics[width=0.19\linewidth]{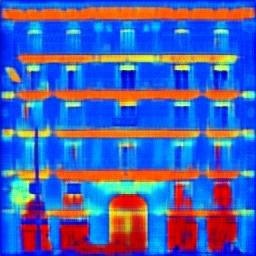}
\includegraphics[width=0.19\linewidth]{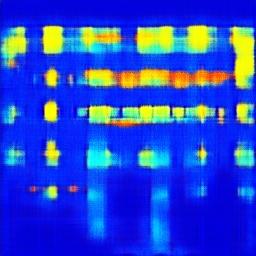}
\includegraphics[width=0.19\linewidth]{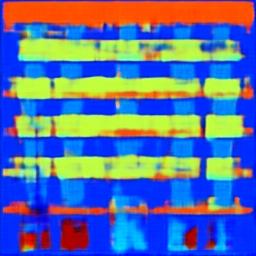}

\includegraphics[width=0.19\linewidth]{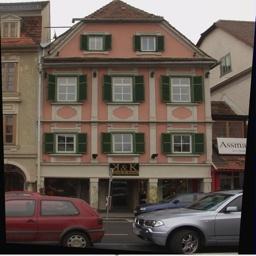}
\includegraphics[width=0.19\linewidth]{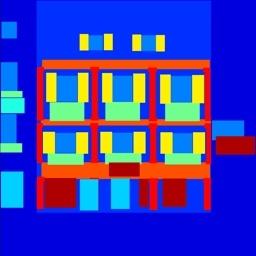}
\includegraphics[width=0.19\linewidth]{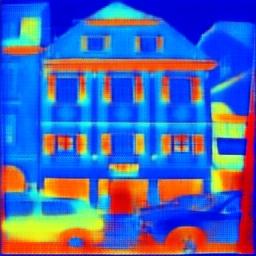}
\includegraphics[width=0.19\linewidth]{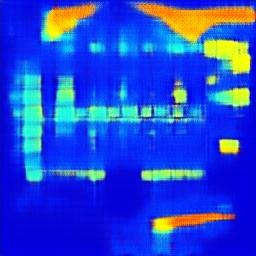}
\includegraphics[width=0.19\linewidth]{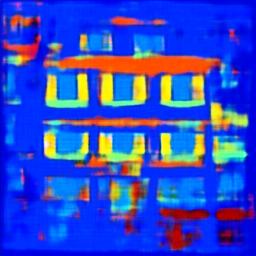}

\includegraphics[width=0.19\linewidth]{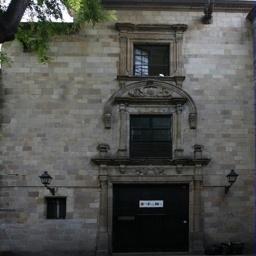}
\includegraphics[width=0.19\linewidth]{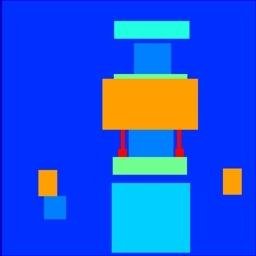}
\includegraphics[width=0.19\linewidth]{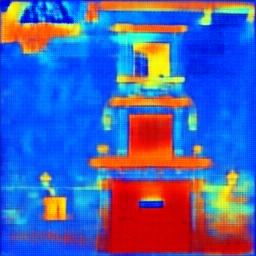}
\includegraphics[width=0.19\linewidth]{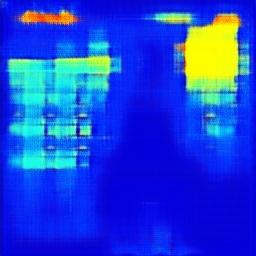}
\includegraphics[width=0.19\linewidth]{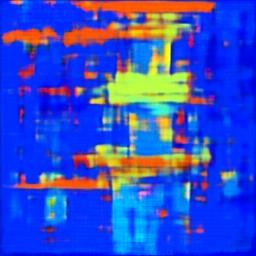}

\begin{subfigure}[]{0.19\linewidth}\caption*{Input}\end{subfigure}
\begin{subfigure}[]{0.19\linewidth}\caption*{GT}\end{subfigure}
\begin{subfigure}[]{0.19\linewidth}\caption*{\textbf{DualGAN}}\end{subfigure}
\begin{subfigure}[]{0.19\linewidth}\caption*{GAN)}\end{subfigure}
\begin{subfigure}[]{0.19\linewidth}\caption*{cGAN~\cite{isola2016image}}\end{subfigure}
\caption{Facades$\rightarrow$label translation. While cGAN correctly labels various 
bulding components such as windows, doors, and balconies, the overall label images 
are not as detailed and structured as DualGAN's outputs.} 
\label{fig:facades2label}
\end{center}
\end{figure}

Using the four labeled datasets, we first compare DualGAN with GAN and cGAN~\cite{isola2016image} on the 
following translation tasks: day$\rightarrow$night (Figure~\ref{fig:day}), labels$\leftrightarrow$facade 
(Figures~\ref{fig:facades} and~\ref{fig:facades2label}), face photo$\leftrightarrow$sketch (Figures~\ref{fig:photo} 
and \ref{fig:sketch}), and map$\leftrightarrow$aerial photo (Figures~\ref{fig:maps} and \ref{fig:aerial}). 
In all these tasks, cGAN was trained with labeled (i.e., paired) data, where we ran the model and code provided 
in~\cite{isola2016image} and chose the optimal loss function for each task: $L_1$ loss for facade$\rightarrow$label 
and $L_1+cGAN$ loss for the other tasks (see~\cite{isola2016image} for more details). In contrast, DualGAN and 
GAN were trained in an unsupervised way, i.e., we decouple the image pairs and then reshuffle the data. The 
results of GAN were generated using our approach by setting $\lambda_U=\lambda_V = 0.0$ in eq.~(\ref{eq:loss_g}), 
noting that this GAN is different from the original GAN model~\cite{goodfellow2014generative} as it employs a conditional 
generator.

All three models were trained on the same training datasets and tested on novel data that does not overlap those for training.
All the training were carried out on a single GeForce GTX Titan X GPU. At test time, all models ran in well under a second 
on this GPU. 

Compared to GAN, in almost all cases, DualGAN produces results that are less blurry, contain fewer artifacts, and better
preserve content structures in the inputs and capture features (e.g., texture, color, and/or style) of the target domain. 
We attribute the improvements to the reconstruction loss, which forces the inputs to be reconstructable from 
outputs through the dual generator and strengthens feedback signals that encodes the targeted distribution.  

In many cases, DualGAN also compares favorably over the supervised cGAN in terms of sharpness of the outputs and 
faithfulness to the input images; see Figures~\ref{fig:day},~\ref{fig:facades},~\ref{fig:photo},~\ref{fig:sketch}, and~\ref{fig:maps}.
This is encouraging since the supervision in cGAN does utilize additional image and pixel correspondences. On the other hand, 
when translating between photos and semantic-based labels, such as map$\leftrightarrow$aerial and label$\leftrightarrow$facades, 
it is often impossible to infer the correspondences between pixel colors and labels based on targeted distribution alone. As a result, 
DualGAN may map pixels to wrong labels (see Figures~\ref{fig:aerial} and~\ref{fig:facades2label}) or labels to wrong colors/textures 
(see Figures~\ref{fig:facades} and~\ref{fig:maps}).

\begin{figure}
\begin{center}
\includegraphics[width=0.32\linewidth]{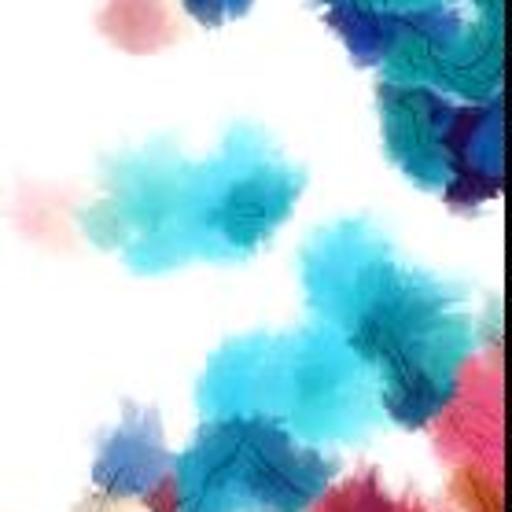}
\includegraphics[width=0.32\linewidth]{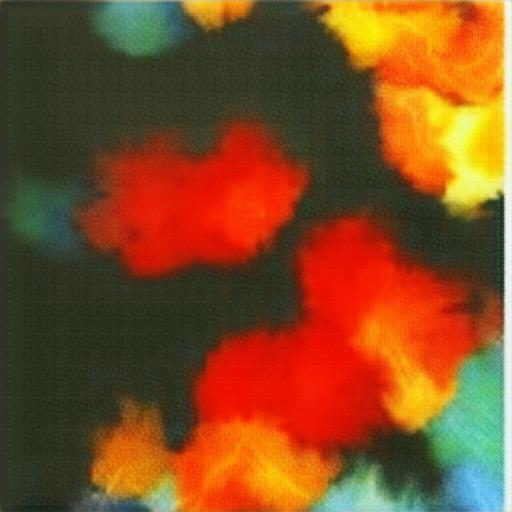}
\includegraphics[width=0.32\linewidth]{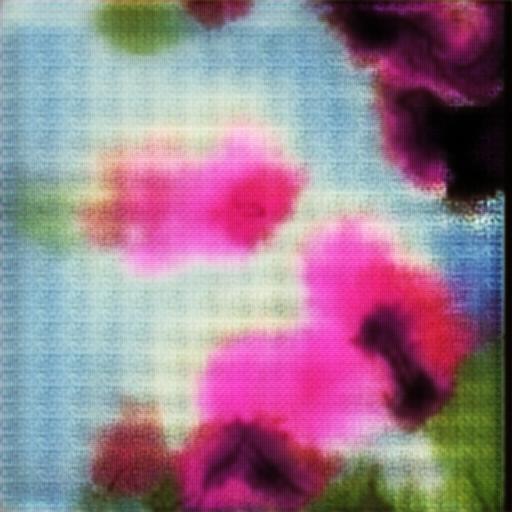}

\includegraphics[width=0.32\linewidth]{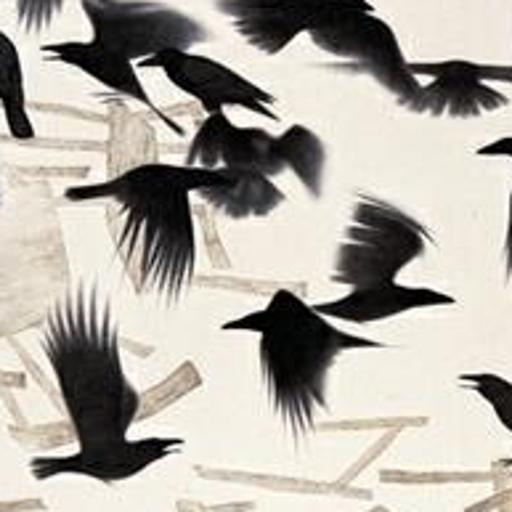}
\includegraphics[width=0.32\linewidth]{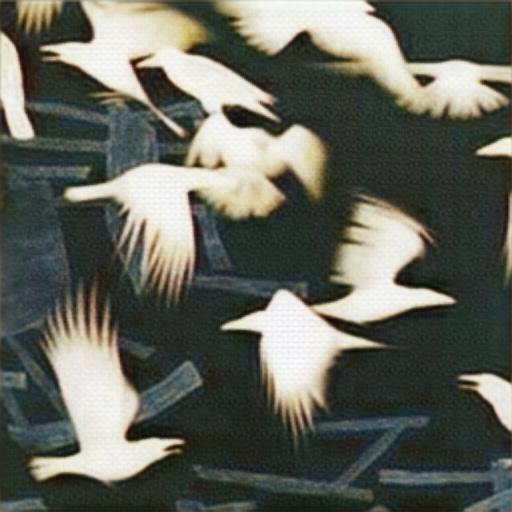}
\includegraphics[width=0.32\linewidth]{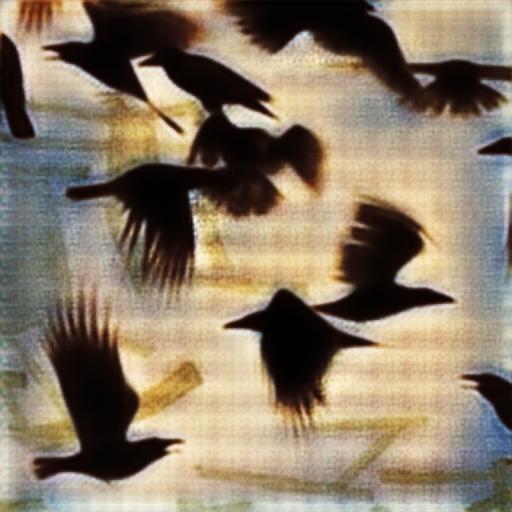}

\includegraphics[width=0.32\linewidth]{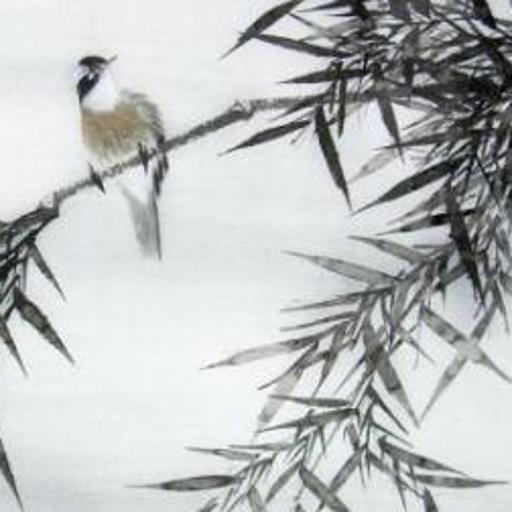}
\includegraphics[width=0.32\linewidth]{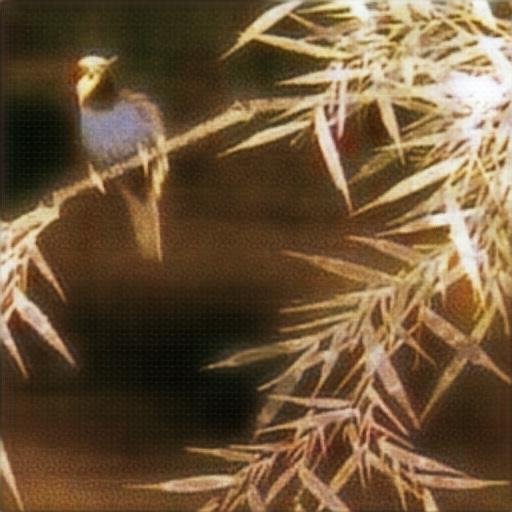}
\includegraphics[width=0.32\linewidth]{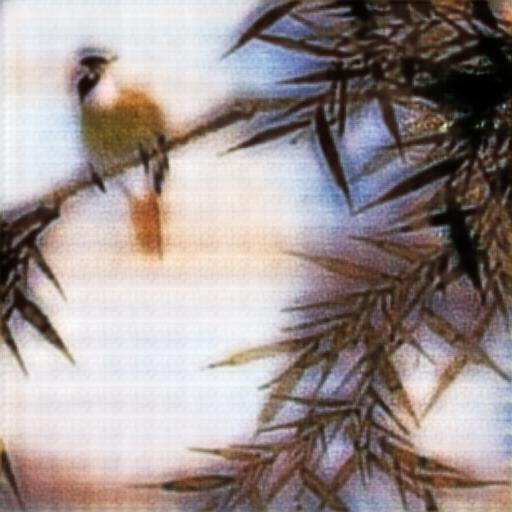}

\includegraphics[width=0.32\linewidth]{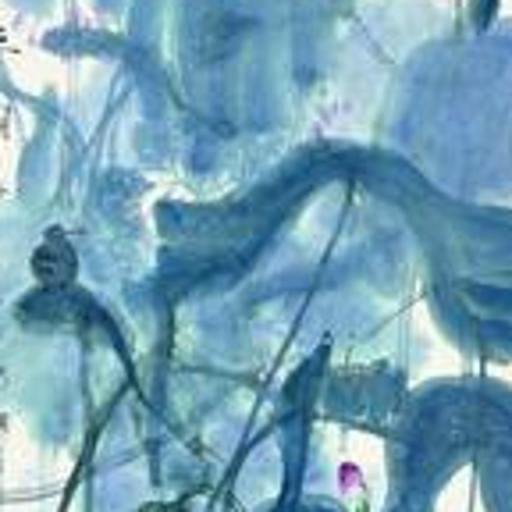}
\includegraphics[width=0.32\linewidth]{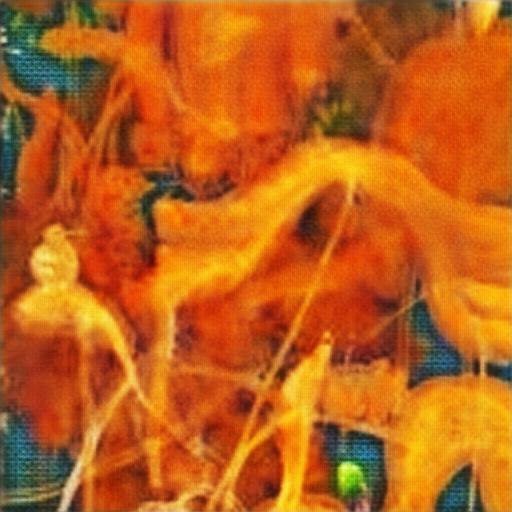}
\includegraphics[width=0.32\linewidth]{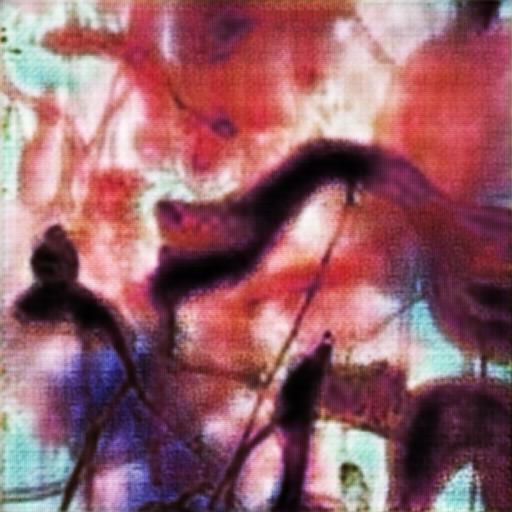}

\begin{subfigure}[]{0.32\linewidth}\caption*{Input}\end{subfigure}
\begin{subfigure}[]{0.32\linewidth}\caption*{\textbf{DualGAN}}\end{subfigure}
\begin{subfigure}[]{0.32\linewidth}\caption*{GAN}\end{subfigure}
\caption{Experimental results for translating Chinese paintings to oil paintings
(without GT available). The background grids shown in the GAN results imply that the outputs of GAN 
are not as stable as those of DualGAN.} \label{fig:chinese}
\end{center}
\end{figure}

\begin{figure*}
\begin{center}

\includegraphics[width=0.13\linewidth]{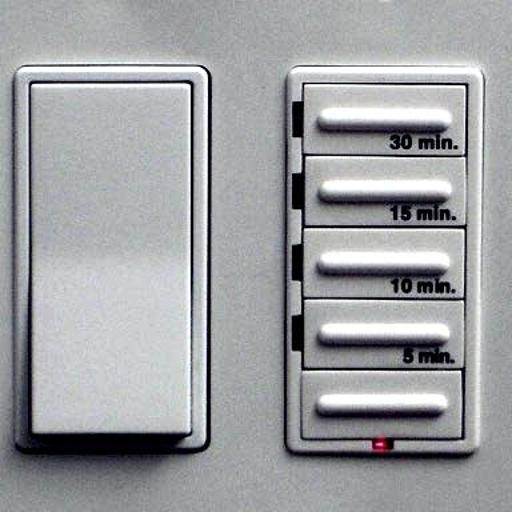}
\includegraphics[width=0.13\linewidth]{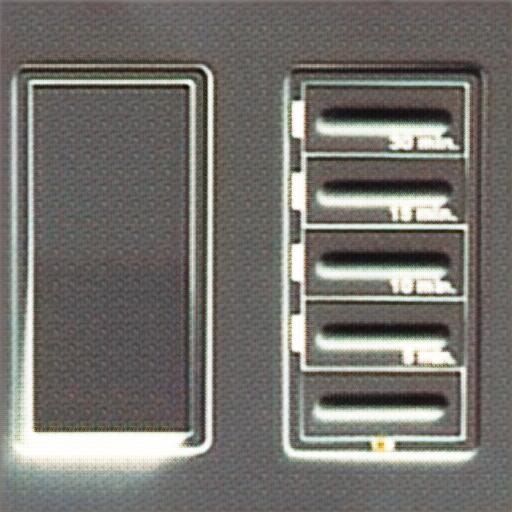}
\includegraphics[width=0.13\linewidth]{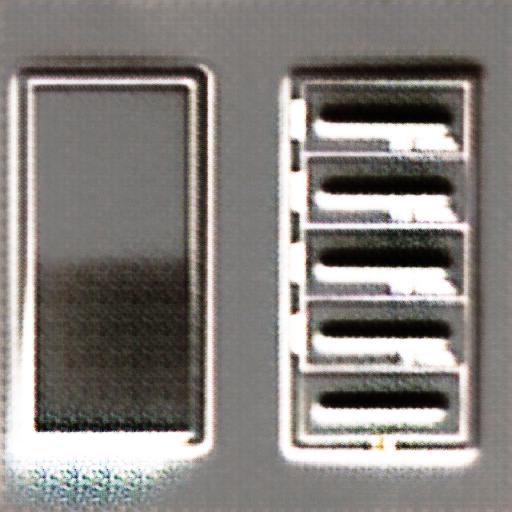}
\includegraphics[width=0.13\linewidth]{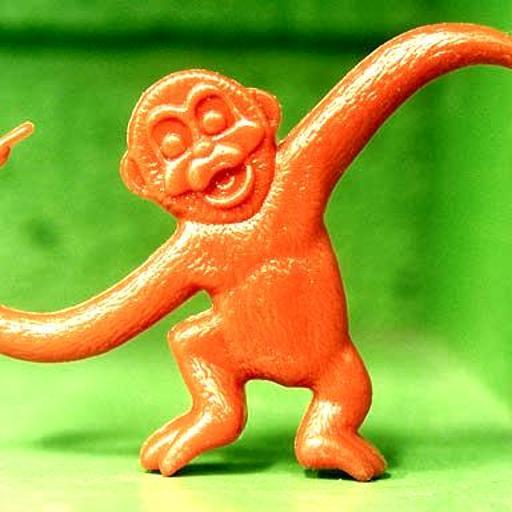}
\includegraphics[width=0.13\linewidth]{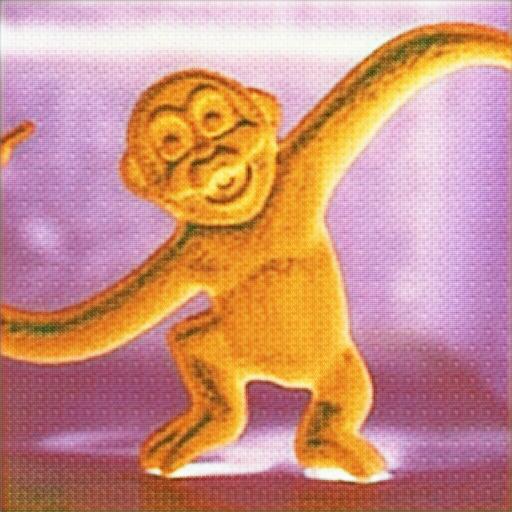}
\includegraphics[width=0.13\linewidth]{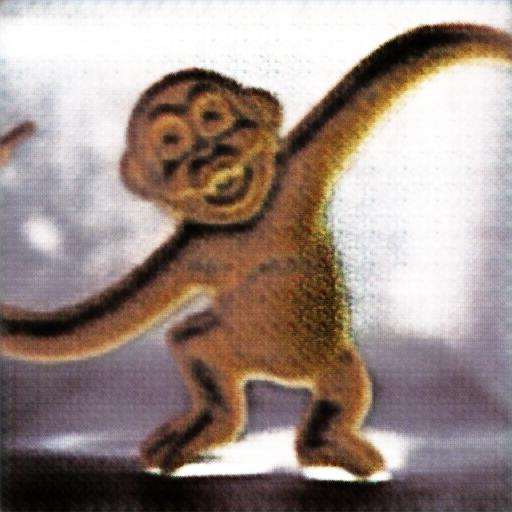}

\begin{subfigure}[]{0.13\linewidth}\caption*{plastic (input)}\end{subfigure}
\begin{subfigure}[]{0.14\linewidth}\caption*{metal (DualGAN)}\end{subfigure}
\begin{subfigure}[]{0.13\linewidth}\caption*{metal (GAN)}\end{subfigure}
\begin{subfigure}[]{0.13\linewidth}\caption*{plastic (input)}\end{subfigure}
\begin{subfigure}[]{0.14\linewidth}\caption*{metal (DualGAN)}\end{subfigure}
\begin{subfigure}[]{0.13\linewidth}\caption*{metal (GAN)}\end{subfigure}

\includegraphics[width=0.13\linewidth]{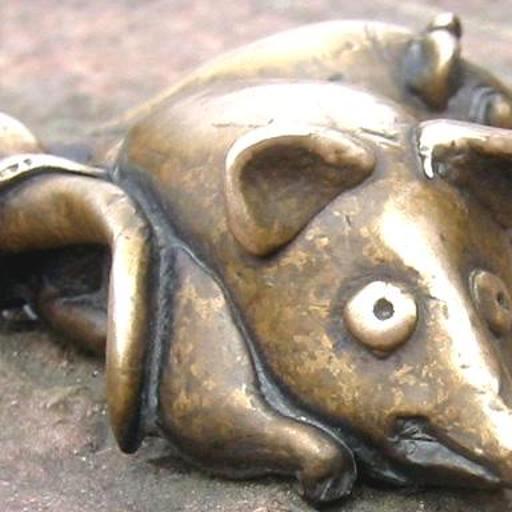}
\includegraphics[width=0.13\linewidth]{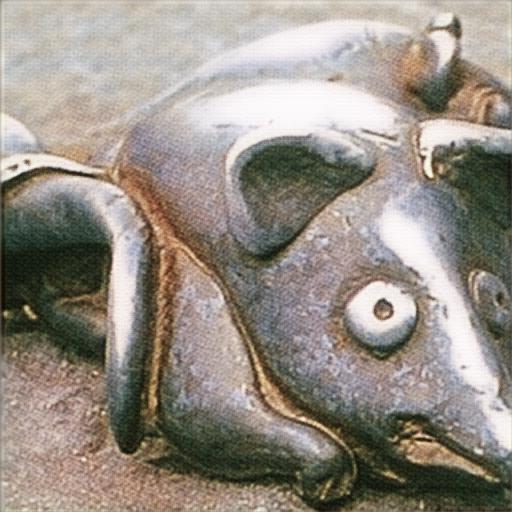}
\includegraphics[width=0.13\linewidth]{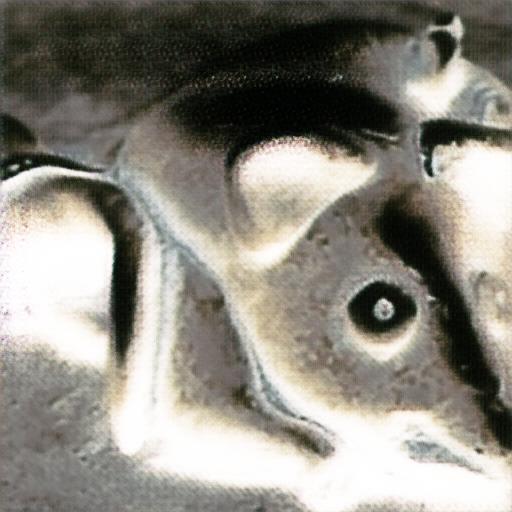}
\includegraphics[width=0.13\linewidth]{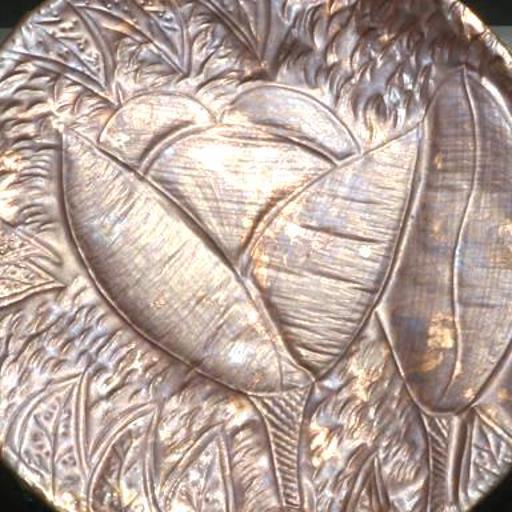}
\includegraphics[width=0.13\linewidth]{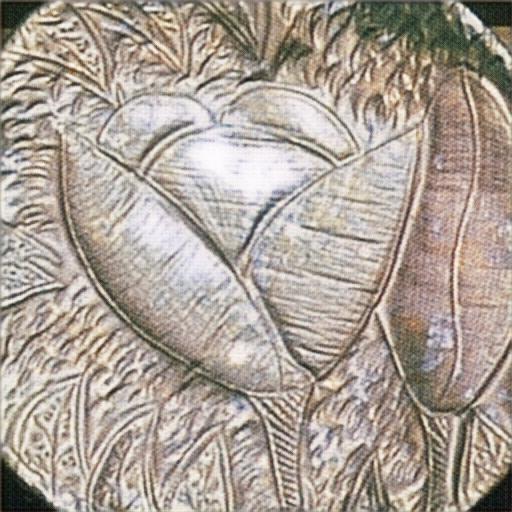}
\includegraphics[width=0.13\linewidth]{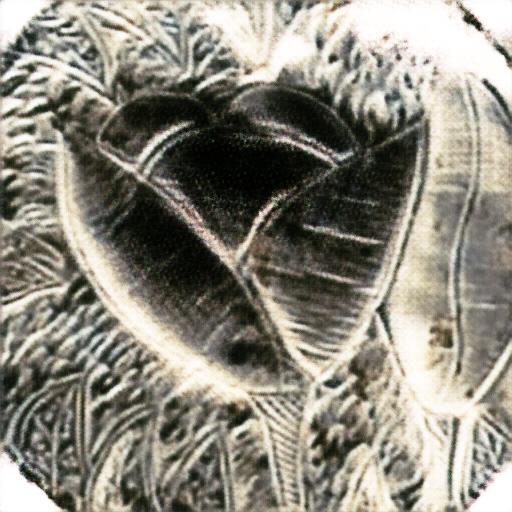}

\begin{subfigure}[]{0.13\linewidth}\caption*{metal (input)}\end{subfigure}
\begin{subfigure}[]{0.13\linewidth}\caption*{stone (DualGAN)}\end{subfigure}
\begin{subfigure}[]{0.13\linewidth}\caption*{stone (GAN)}\end{subfigure}
\begin{subfigure}[]{0.13\linewidth}\caption*{metal (input)}\end{subfigure}
\begin{subfigure}[]{0.13\linewidth}\caption*{stone (DualGAN)}\end{subfigure}
\begin{subfigure}[]{0.13\linewidth}\caption*{stone (GAN)}\end{subfigure}

\includegraphics[width=0.13\linewidth]{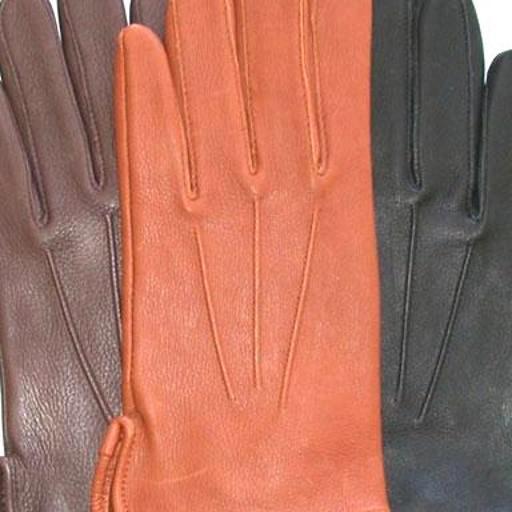}
\includegraphics[width=0.13\linewidth]{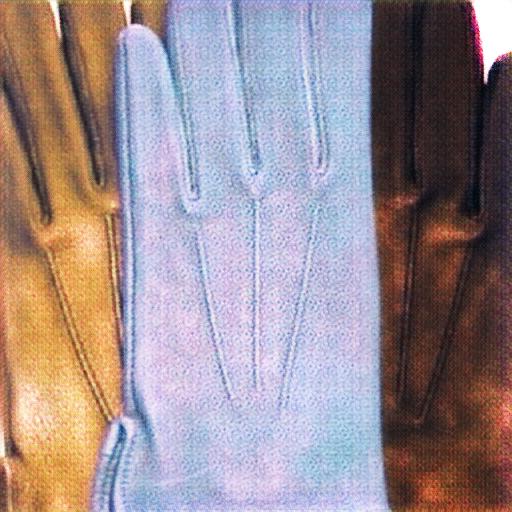}
\includegraphics[width=0.13\linewidth]{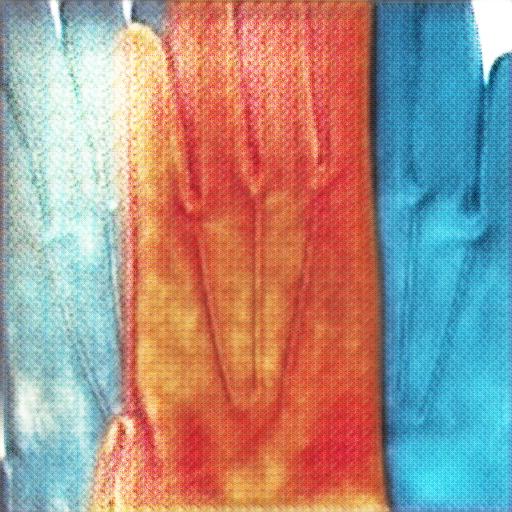}
\includegraphics[width=0.13\linewidth]{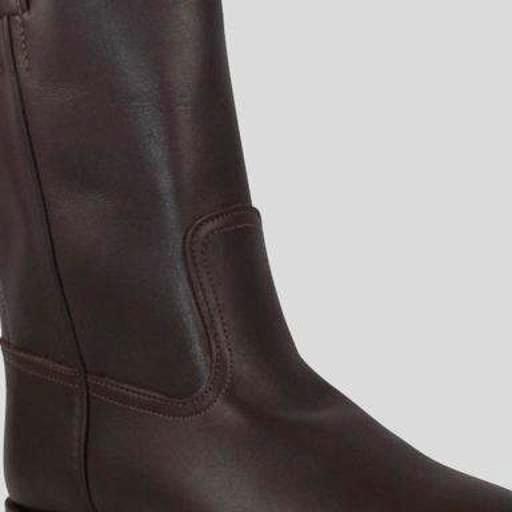}
\includegraphics[width=0.13\linewidth]{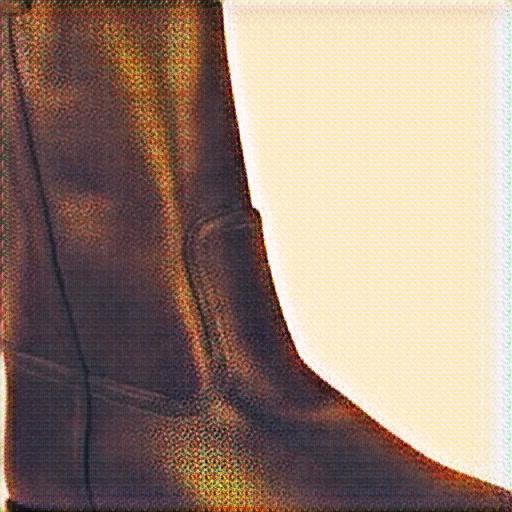}
\includegraphics[width=0.13\linewidth]{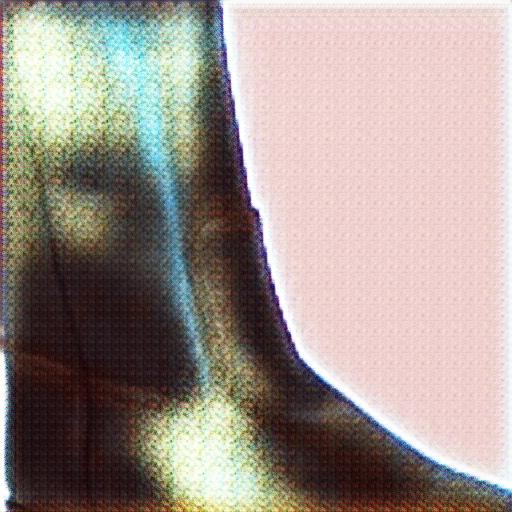}

\begin{subfigure}[]{0.13\linewidth}\caption*{leather (input)}\end{subfigure}
\begin{subfigure}[]{0.14\linewidth}\caption*{fabric (DualGAN)}\end{subfigure}
\begin{subfigure}[]{0.13\linewidth}\caption*{fabric (GAN)}\end{subfigure}
\begin{subfigure}[]{0.13\linewidth}\caption*{leather (input)}\end{subfigure}
\begin{subfigure}[]{0.14\linewidth}\caption*{fabric (DualGAN)}\end{subfigure}
\begin{subfigure}[]{0.13\linewidth}\caption*{fabric (GAN)}\end{subfigure}

\includegraphics[width=0.13\linewidth]{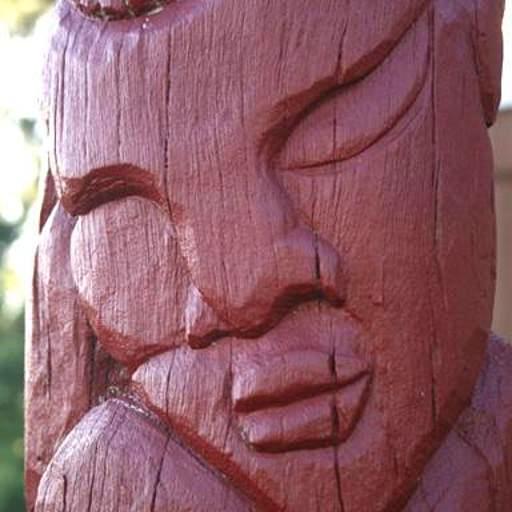}
\includegraphics[width=0.13\linewidth]{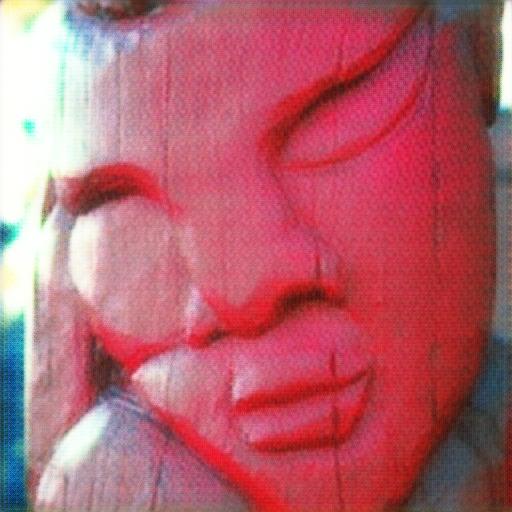}
\includegraphics[width=0.13\linewidth]{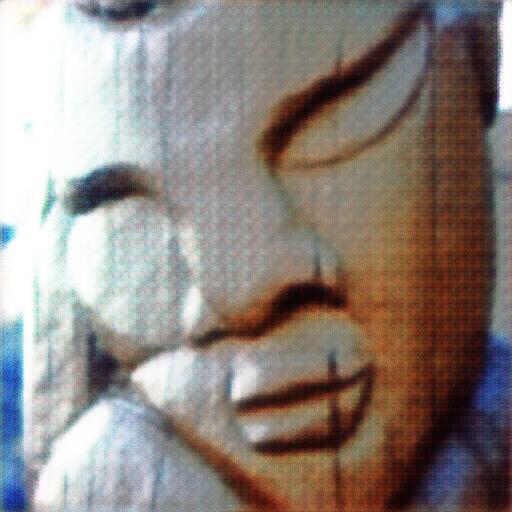}
\includegraphics[width=0.13\linewidth]{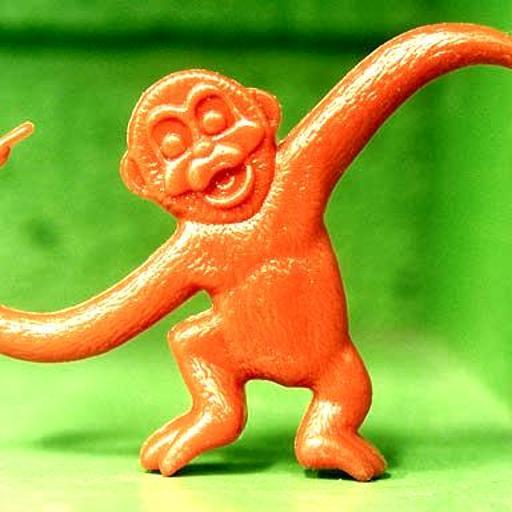}
\includegraphics[width=0.13\linewidth]{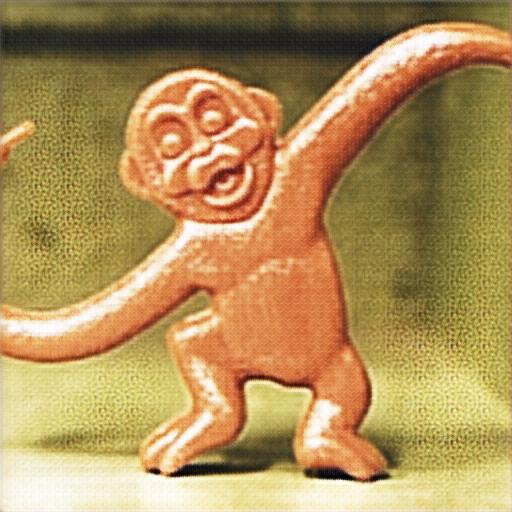}
\includegraphics[width=0.13\linewidth]{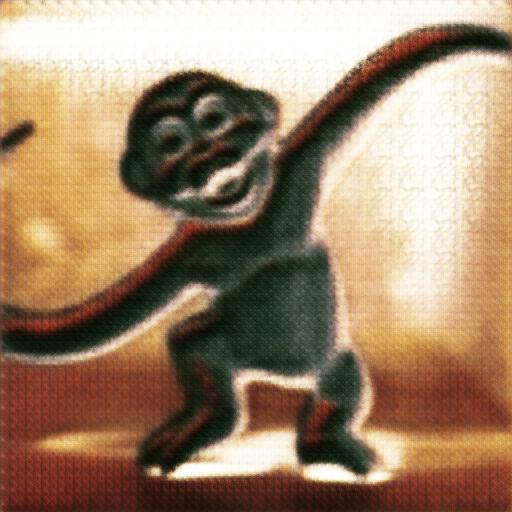}

\begin{subfigure}[]{0.13\linewidth}\caption*{wood (input)}\end{subfigure}
\begin{subfigure}[]{0.15\linewidth}\caption*{plastic (DualGAN)}\end{subfigure}
\begin{subfigure}[]{0.13\linewidth}\caption*{plastic (GAN)}\end{subfigure}
\begin{subfigure}[]{0.13\linewidth}\caption*{plastic (input)}\end{subfigure}
\begin{subfigure}[]{0.14\linewidth}\caption*{wood (DualGAN)}\end{subfigure}
\begin{subfigure}[]{0.13\linewidth}\caption*{wood (GAN)}\end{subfigure}
\caption{Experimental results for various material transfer tasks. From top to bottom, 
plastic$\rightarrow$metal, metal$\rightarrow$stone, leather$\rightarrow$fabric, and 
plastic$\leftrightarrow$wood.} \label{fig:material}
\end{center}
\end{figure*}

Figures~\ref{fig:chinese} and \ref{fig:material} show image translation results obtained using the two unlabeled datasets, including 
oil$\leftrightarrow$Chinese, plastic$\rightarrow$metal, metal$\rightarrow$stone, leather$\rightarrow$fabric, as well as wood$\leftrightarrow$plastic. 
The results demonstrate that visually convincing images can be generated by DualGAN when no corresponding images can be found in 
the target domains. As well, the DualGAN results generally contain less artifacts than those from GAN.


%
%

\subsection{Quantitative evaluation}

To quantitatively evaluate DualGAN, we set up two user studies through Amazon Mechanical Turk (AMT). 
The ``material perceptual'' test evaluates the material transfer results, in which we mix the outputs 
from all material transfer tasks and let the Turkers choose the best match based on which material they 
believe the objects in the image are made of. For a total of 176 output images, each was evaluated by 
ten Turkers. An output image is rated as a success if at least three Turkers selected the target material type. 
Success rates of various material transfer results using different approaches are summarized in 
Table~\ref{table:material}, showing that DualGAN outperforms GAN by a large margin.

In addition, we run the AMT ``realness score'' evaluation for sketch$\rightarrow$photo, label map$\rightarrow$facades, 
maps$\rightarrow$aerial photo, and day$\rightarrow$night translations. To eliminate potential bias, for each of the four 
evaluations, we randomly shuffle real photos and outputs from all three approaches before showing them to Turkers.  
Each image is shown to 20 Turkers, who were asked to score the image based on to what extent the synthesized photo 
looks real. The ``realness'' score ranges from 0 (totally missing), 1 (bad), 2 (acceptable), 3 (good), to 4 (compelling). The 
average score of different approaches on various tasks are then computed and shown in Table.~\ref{table:score}. The 
AMT study results show that DualGAN outperforms GAN on all tasks and outperforms cGAN on two tasks as well. This 
indicates that cGAN has little tolerance to misalignment and inconsistency between image pairs, but the additional 
pixel-level correspondence does help cGAN correctly map labels to colors and textures.


Finally, we compute the segmentation accuracies for facades$\rightarrow$label and aerial$\rightarrow$map tasks, as reported
in Tables~\ref{table:acc} and \ref{table:acc_maps}. The comparison shows that DualGAN is outperformed by cGAN, which is 
expected as it is difficult to infer proper labeling without image correspondence information from the training data.

\begin{table}
\begin{center}
\begin{tabular}{|c|c|c|}
\hline
Task  & DualGAN  &  GAN \\
\hline\hline
plastic$\rightarrow$wood &  \textbf{2}/11 & 0/11 \\
\hline
wood$\rightarrow$plastic &  \textbf{1}/11 & 0/11 \\
\hline
metal$\rightarrow$stone&   \textbf{2}/11 & 0/11 \\
\hline
stone$\rightarrow$metal &   \textbf{2}/11 & 0/11 \\
\hline
leather$\rightarrow$fabric &  \textbf{3}/11 & 2/11 \\
\hline
fabric$\rightarrow$leather &  \textbf{2}/11 & 1/11 \\
\hline
plastic$\rightarrow$metal &   \textbf{7}/11 & 3/11 \\
\hline
metal$\rightarrow$plastic &   \textbf{1}/11 & 0/11 \\
\hline
\end{tabular}
\caption{Success rates of various material transfer tasks based on the AMT ``material perceptual'' test. 
There are 11 images in each set of transfer result, with noticeable improvements of DualGAN over GAN.} 
\label{table:material}
\end{center}
\end{table}

\begin{table}
\tabcolsep=0.11cm
\begin{center}
\begin{tabular}{c|cccc}
\hline
	   &   \multicolumn{4}{|c}{Avg. ``realness'' score }\\
		  Task & DualGAN  &  cGAN\cite{isola2016image} & GAN & GT\\
\hline\hline

sketch$\rightarrow$photo & \textbf{1.87 }&  1.69   & 1.04  & 3.56 \\
\hline
day$\rightarrow$night &  \textbf{2.42 } & 1.89   &  0.13  & 3.05  \\
\hline
label$\rightarrow$facades & 1.89  &   \textbf{2.59 } & 1.43  & 3.33 \\
\hline
map$\rightarrow$aerial & 2.52 &  \textbf{2.92 } & 1.88  &  3.21  \\
\hline
\end{tabular}
\caption{Average AMT ``realness'' scores of outputs from various tasks. The results show that 
DualGAN outperforms GAN in all tasks. It also outperforms cGAN for sketch$\rightarrow$photo 
and day$\rightarrow$night tasks, but still lag behind for label$\rightarrow$facade and 
map$\rightarrow$aerial tasks. In the latter two tasks, the additional image correspondence in 
training data would help cGAN map labels to the proper colors/textures.}
\label{table:score}
\end{center}
\end{table}

\tabcolsep=0.11cm
\begin{table}
\begin{center}
\begin{tabular}{|c|c|c|c|}
\hline
  & Per-pixel acc.  & Per-class acc. & Class IOU\\
\hline\hline
DualGAN &  0.27 &  0.13 & 0.06\\      
\hline
cGAN~\cite{isola2016image}  &  \textbf{0.54}&  \textbf{0.33} &\textbf{0.19}\\  
\hline
GAN &   0.22&   0.10 & 0.05 \\
\hline
\end{tabular}
\caption{Segmentation accuracy for the facades$\rightarrow$label task. DualGAN outperforms GAN, but is not as accurate as cGAN. 
Without image correspondence (for cGAN), even if DualGAN segments a region properly, it may not assign the region with a 
correct label.} \label{table:acc}
\end{center}
\end{table}

\tabcolsep=0.11cm
\begin{table}
\begin{center}
\begin{tabular}{|c|c|c|c|}
\hline
  & Per-pixel acc.  & Per-class acc. & Class IOU\\
\hline\hline
DualGAN &  0.42 &  0.22 & 0.09\\      
\hline
cGAN~\cite{isola2016image}  &  \textbf{0.70}&  \textbf{0.46} &\textbf{0.26}\\  
\hline
GAN &   0.41&   0.23 & 0.09 \\
\hline
\end{tabular}
\caption{Segmentation accuracy for the aerial$\rightarrow$map task, for which DualGAN performs less than 
satisfactorily.} \label{table:acc_maps}
\end{center}
\end{table}

\section{Conclusion}

We propose DualGAN, a novel unsupervised dual learning framework for general-purpose image-to-image translation. The 
unsupervised characteristic of DualGAN enables many real world applications, as demonstrated in this work, as well as
in the concurrent work CycleGAN~\cite{zhu2017CycleGAN}. Experimental results suggest that the DualGAN mechanism 
can significantly improve the outputs of GAN for various image-to-image translation tasks. With unlabeled data only, 
DualGAN can generate comparable or even better outputs than conditional GAN~\cite{isola2016image} which is trained 
with labeled data providing image and pixel-level correspondences.

On the other hand, our method is outperformed by conditional GAN or cGAN~\cite{isola2016image} for certain tasks
which involve semantics-based labels. This is due to the lack of pixel and label correspondence information, which 
cannot be inferred from the target distribution alone. In the future, we intend to investigate whether this limitation can 
be lifted with the use of a small number of labeled data as a warm start.

\paragraph{Acknowledgment.}
We thank all the anonymous reviewers for their valuable comments and suggestions. The first author is a PhD student from the Memorial University 
of Newfoundland and has been visiting SFU since 2016. This work was supported in part by grants from the Natural Sciences and Engineering 
Research Council (NSERC) of Canada (No.~611370, 2017-06086).

\newpage

\bibliographystyle{ieee}
\bibliography{bib}

\begin{figure*}
\begin{center}
\includegraphics[width=0.13\linewidth]{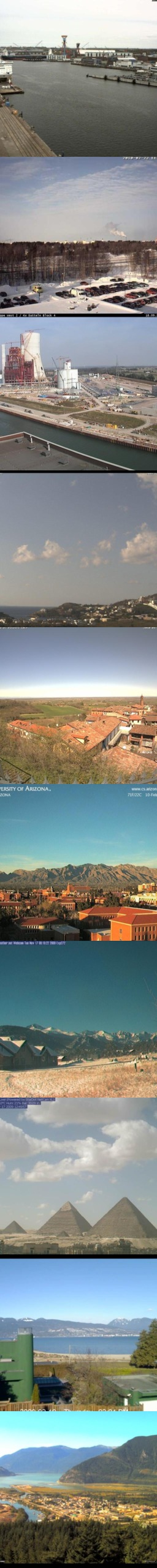}
\includegraphics[width=0.13\linewidth]{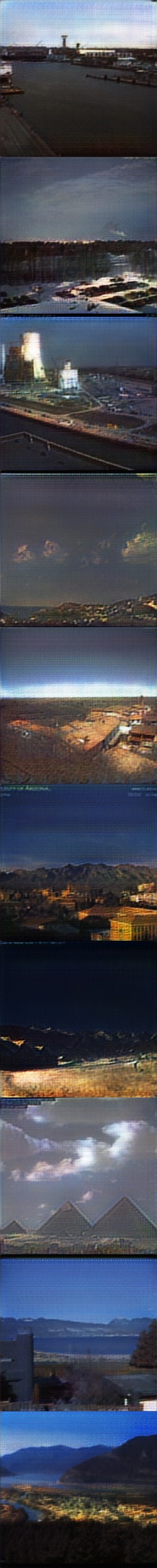}
\includegraphics[width=0.13\linewidth]{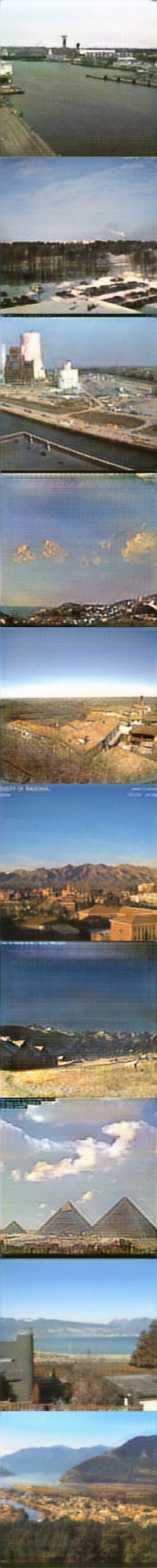}
\includegraphics[width=0.13\linewidth]{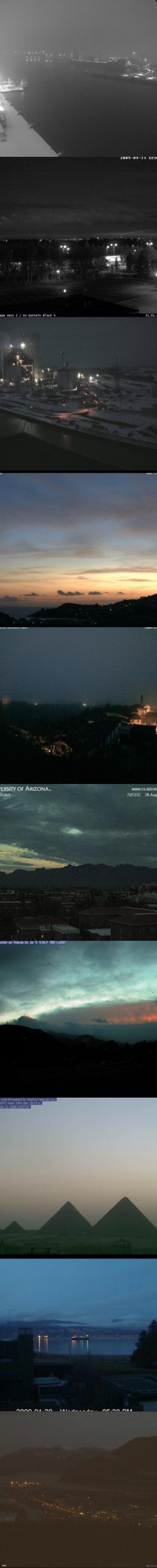}
\includegraphics[width=0.13\linewidth]{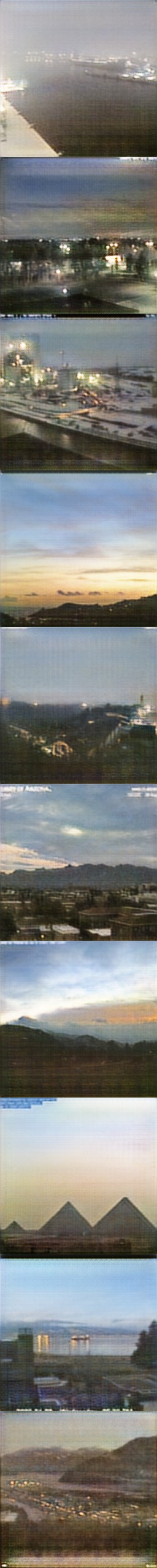}
\includegraphics[width=0.13\linewidth]{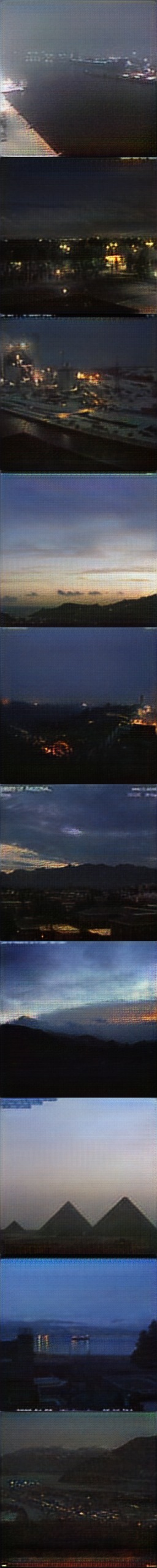}

\begin{subfigure}[]{0.13\linewidth}\caption*{$U$ (day)}\end{subfigure}
\begin{subfigure}[]{0.13\linewidth}\caption*{$G_A(U)$ (night)}\end{subfigure}
\begin{subfigure}[]{0.13\linewidth}\caption*{$G_B(G_A(U))$ (day)}\end{subfigure}
\begin{subfigure}[]{0.13\linewidth}\caption*{$V$ (night)}\end{subfigure}
\begin{subfigure}[]{0.13\linewidth}\caption*{$G_B(V)$ (day)}\end{subfigure}
\begin{subfigure}[]{0.13\linewidth}\caption*{$G_A(G_B(V))$ (night)}\end{subfigure}
\caption{day scenes$\rightarrow$night scenes translation results by DualGAN} 
\label{fig:da2ni}
\end{center}
\end{figure*}

\begin{figure*}
\begin{center}
\includegraphics[width=0.13\linewidth]{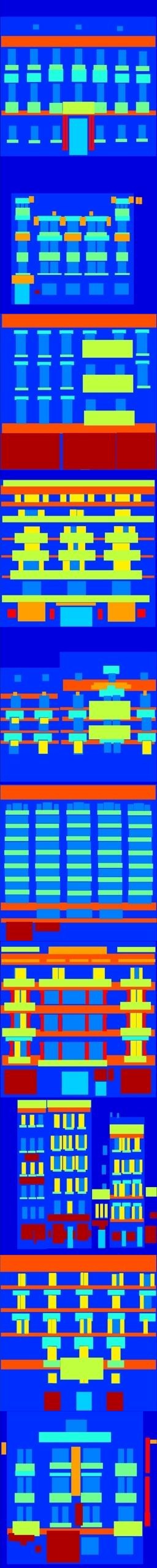}
\includegraphics[width=0.13\linewidth]{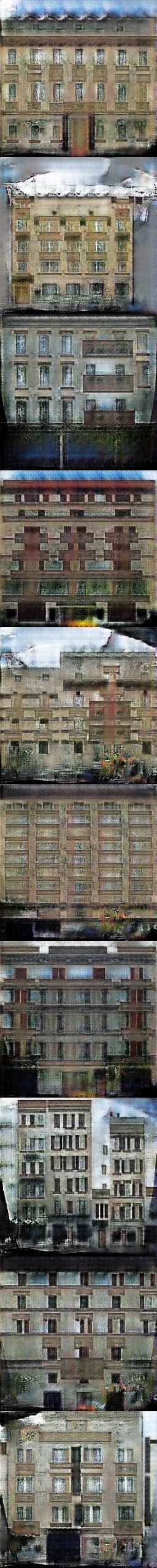}
\includegraphics[width=0.13\linewidth]{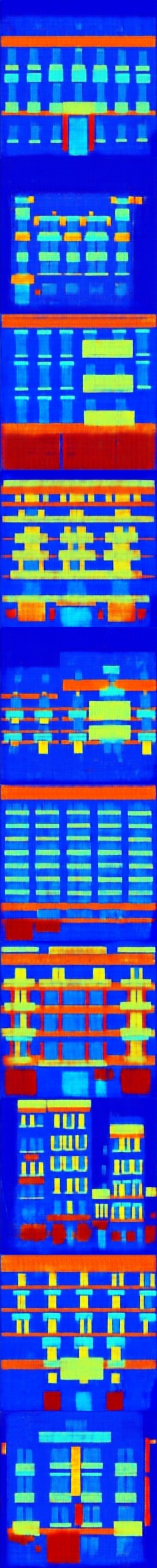}
\includegraphics[width=0.13\linewidth]{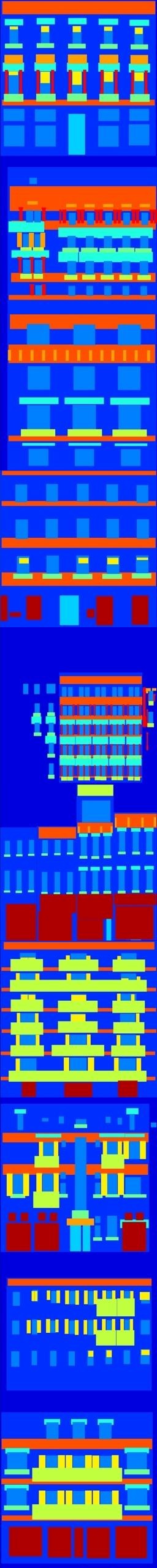}
\includegraphics[width=0.13\linewidth]{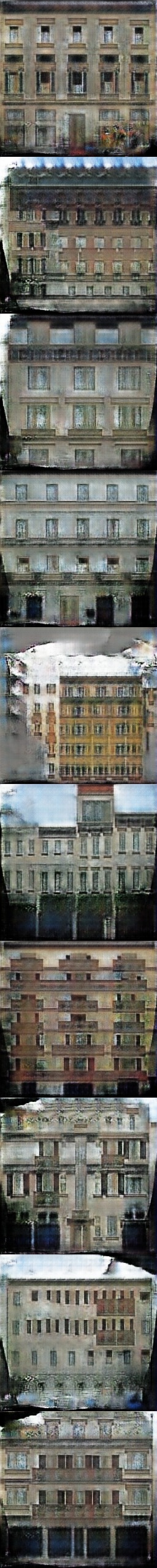}
\includegraphics[width=0.13\linewidth]{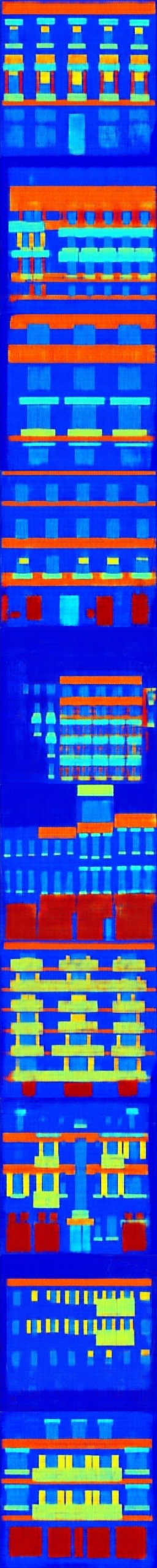}

\begin{subfigure}[]{0.13\linewidth}\caption*{$V$ (label)}\end{subfigure}
\begin{subfigure}[]{0.13\linewidth}\caption*{$G_B(V)$ (photo)}\end{subfigure}
\begin{subfigure}[]{0.13\linewidth}\caption*{$G_A(G_B(V))$ (label)}\end{subfigure}
\begin{subfigure}[]{0.13\linewidth}\caption*{$V$ (label)}\end{subfigure}
\begin{subfigure}[]{0.13\linewidth}\caption*{$G_B(V)$ (photo)}\end{subfigure}
\begin{subfigure}[]{0.13\linewidth}\caption*{$G_A(G_B(V))$ (label)}\end{subfigure}
\caption{label map$\rightarrow$photo translation results by DualGAN} 
\label{fig:la2ph}
\end{center}
\end{figure*}

\begin{figure*}
\begin{center}
\includegraphics[width=0.13\linewidth]{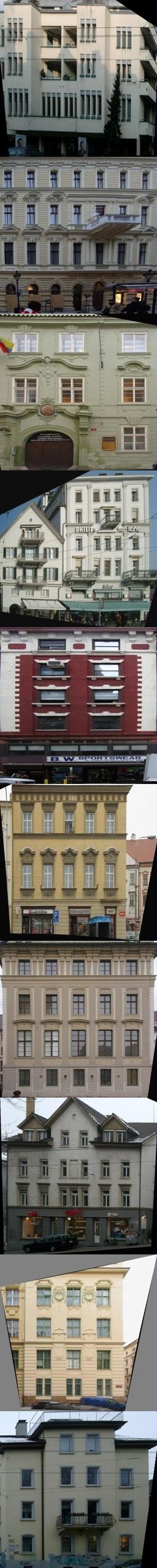}
\includegraphics[width=0.13\linewidth]{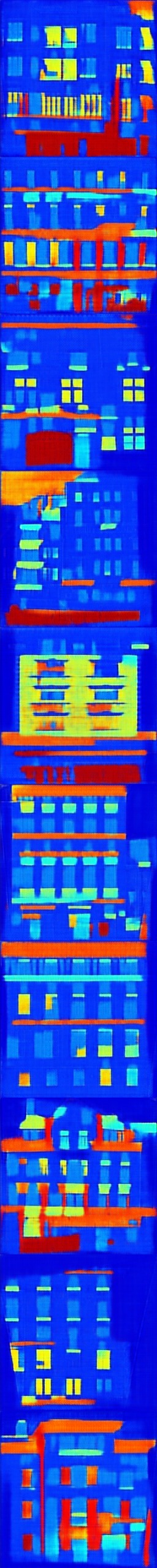}
\includegraphics[width=0.13\linewidth]{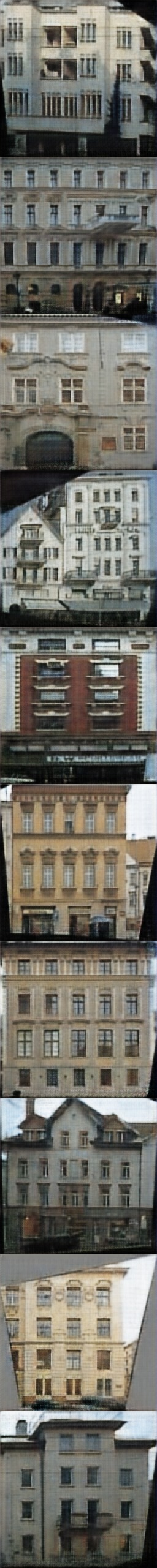}
\includegraphics[width=0.13\linewidth]{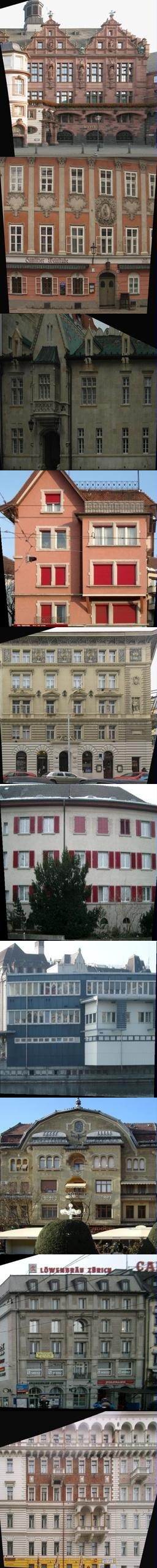}
\includegraphics[width=0.13\linewidth]{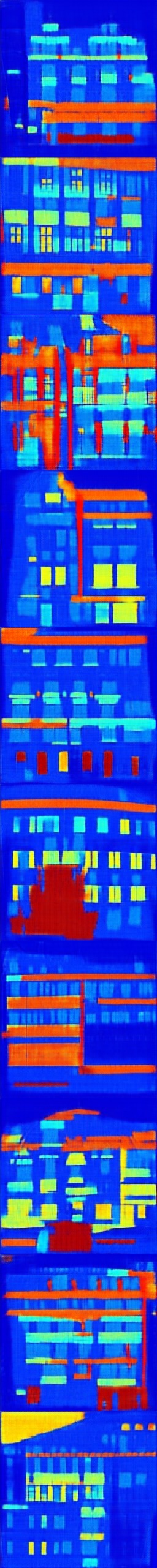}
\includegraphics[width=0.13\linewidth]{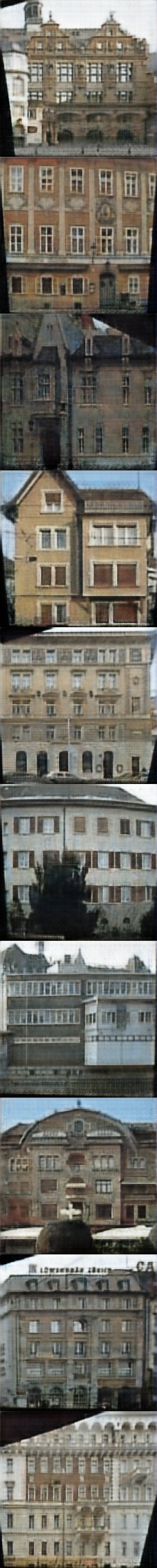}

\begin{subfigure}[]{0.13\linewidth}\caption*{$U$ (photo)}\end{subfigure}
\begin{subfigure}[]{0.13\linewidth}\caption*{$G_A(U)$ (label)}\end{subfigure}
\begin{subfigure}[]{0.13\linewidth}\caption*{$G_B(G_A(U))$ (photo)}\end{subfigure}
\begin{subfigure}[]{0.13\linewidth}\caption*{$U$ (photo)}\end{subfigure}
\begin{subfigure}[]{0.13\linewidth}\caption*{$G_A(U)$ (label)}\end{subfigure}
\begin{subfigure}[]{0.13\linewidth}\caption*{$G_B(G_A(U))$ (photo)}\end{subfigure}
\caption{photo$\rightarrow$label map translation results by DualGAN} 
\label{fig:ph2la}
\end{center}
\end{figure*}

\begin{figure*}
\begin{center}
\includegraphics[width=0.13\linewidth]{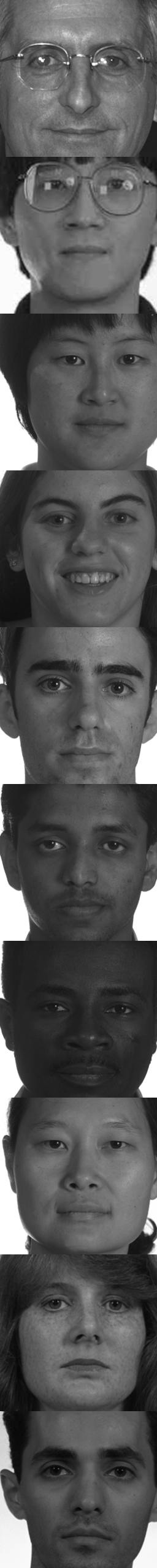}
\includegraphics[width=0.13\linewidth]{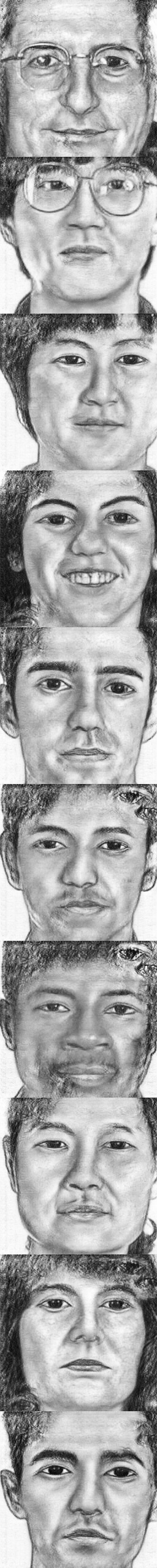}
\includegraphics[width=0.13\linewidth]{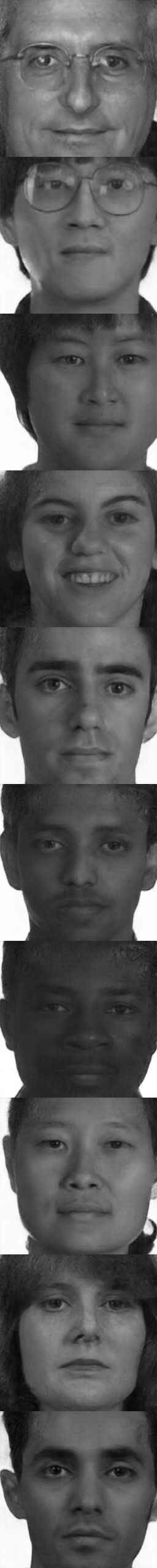}
\includegraphics[width=0.13\linewidth]{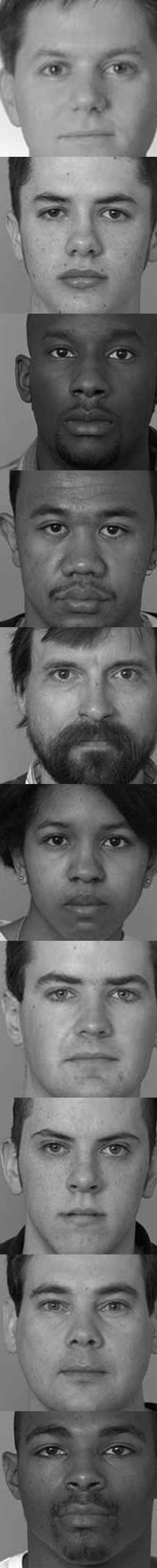}
\includegraphics[width=0.13\linewidth]{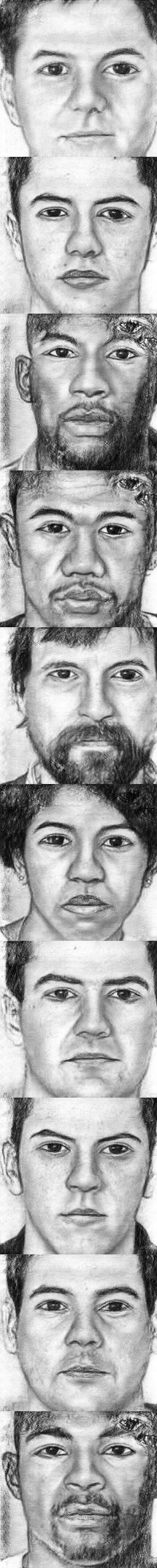}
\includegraphics[width=0.13\linewidth]{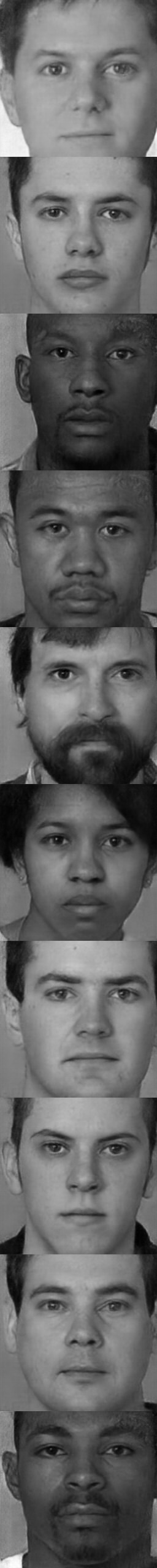}

\begin{subfigure}[]{0.13\linewidth}\caption*{$V$ (photo)}\end{subfigure}
\begin{subfigure}[]{0.13\linewidth}\caption*{$G_B(V)$ (sketch)}\end{subfigure}
\begin{subfigure}[]{0.13\linewidth}\caption*{$G_A(G_B(V))$ (photo)}\end{subfigure}
\begin{subfigure}[]{0.13\linewidth}\caption*{$V$ (photo)}\end{subfigure}
\begin{subfigure}[]{0.13\linewidth}\caption*{$G_B(V)$ (sketch)}\end{subfigure}
\begin{subfigure}[]{0.13\linewidth}\caption*{$G_A(G_B(V))$ (photo)}\end{subfigure}
\caption{Photo$\rightarrow$sketch translation results by DualGAN} 
\label{fig:ph2sk}
\end{center}
\end{figure*}

\begin{figure*}
\begin{center}
\includegraphics[width=0.13\linewidth]{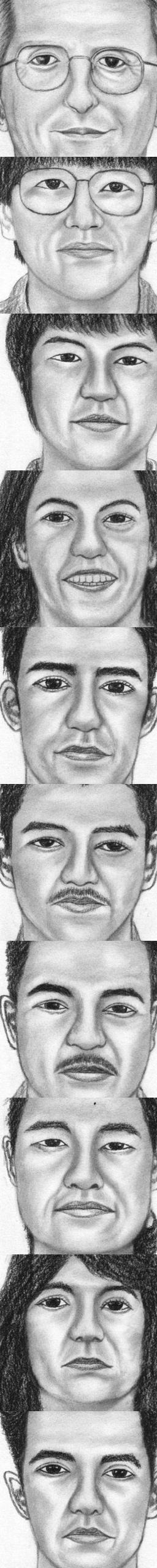}
\includegraphics[width=0.13\linewidth]{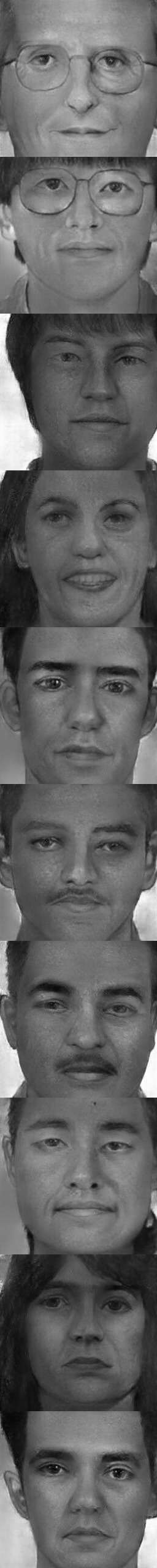}
\includegraphics[width=0.13\linewidth]{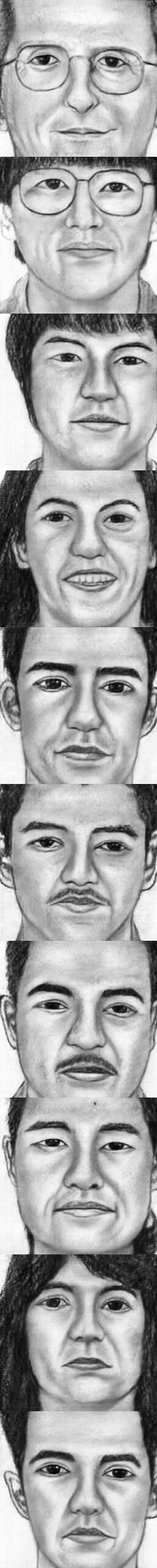}
\includegraphics[width=0.13\linewidth]{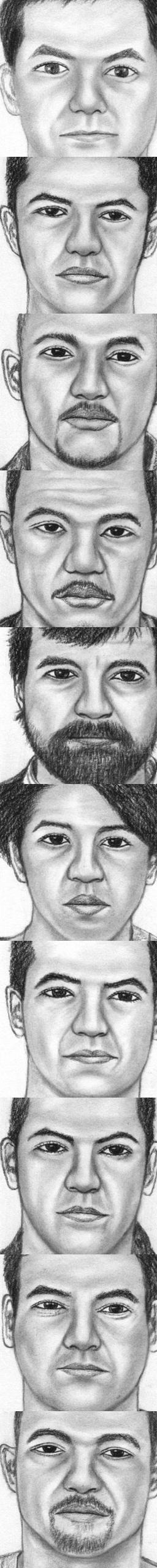}
\includegraphics[width=0.13\linewidth]{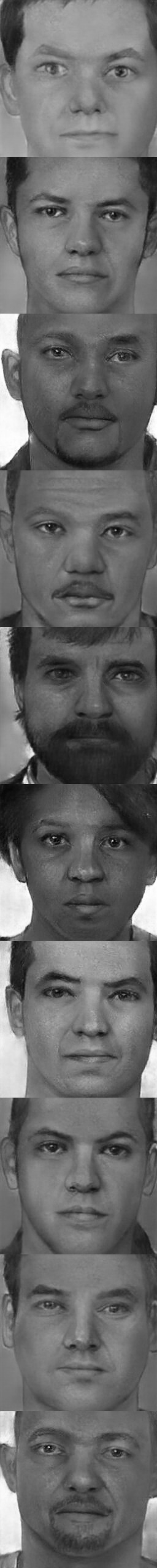}
\includegraphics[width=0.13\linewidth]{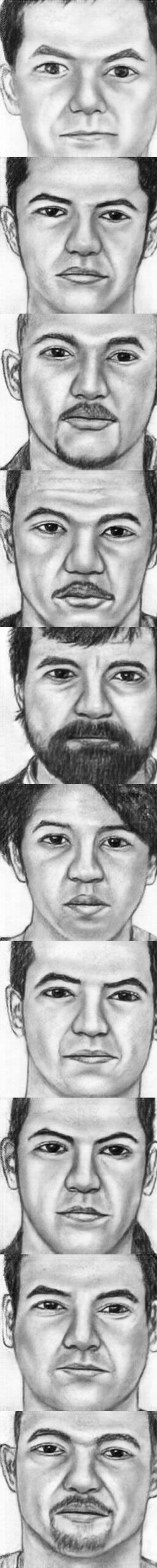}

\begin{subfigure}[]{0.13\linewidth}\caption*{$U$ (sketch)}\end{subfigure}
\begin{subfigure}[]{0.13\linewidth}\caption*{$G_A(U)$ (photo)}\end{subfigure}
\begin{subfigure}[]{0.13\linewidth}\caption*{$G_B(G_A(U))$ (sketch)}\end{subfigure}
\begin{subfigure}[]{0.13\linewidth}\caption*{$U$ (sketch)}\end{subfigure}
\begin{subfigure}[]{0.13\linewidth}\caption*{$G_A(U)$ (photo)}\end{subfigure}
\begin{subfigure}[]{0.13\linewidth}\caption*{$G_B(G_A(U))$ (sketch)}\end{subfigure}
\caption{sketch$\rightarrow$photo translation results by DualGAN} 
\label{fig:sk2ph}
\end{center}
\end{figure*}

\begin{figure*}
\begin{center}
\includegraphics[width=0.13\linewidth]{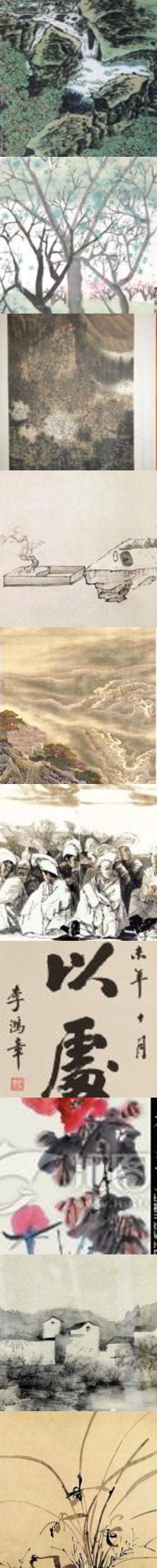}
\includegraphics[width=0.13\linewidth]{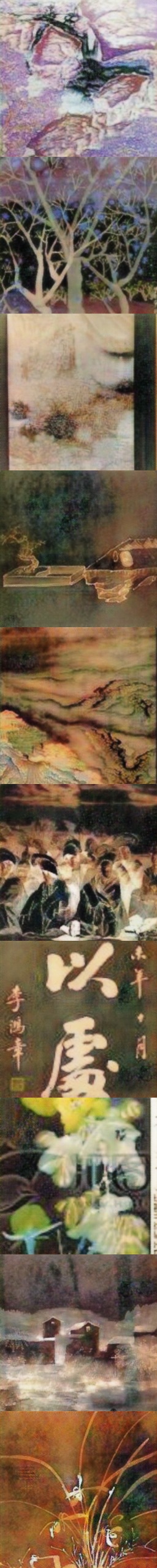}
\includegraphics[width=0.13\linewidth]{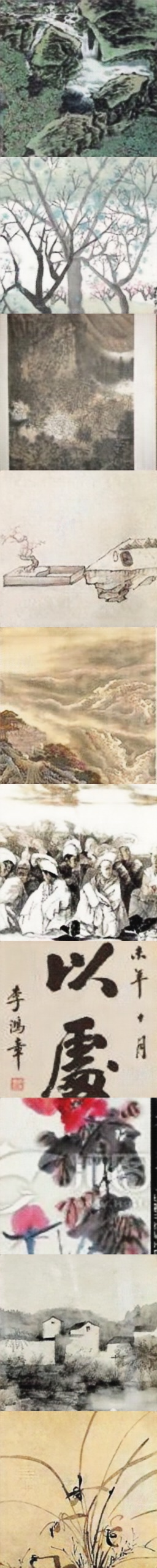}
\includegraphics[width=0.13\linewidth]{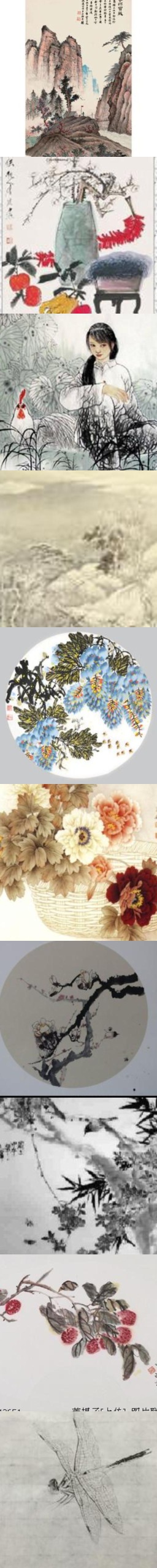}
\includegraphics[width=0.13\linewidth]{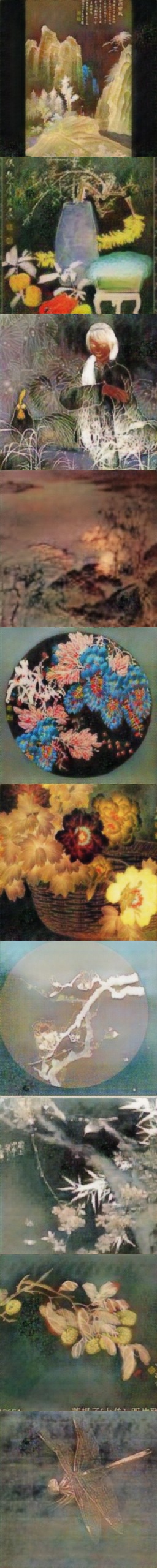}
\includegraphics[width=0.13\linewidth]{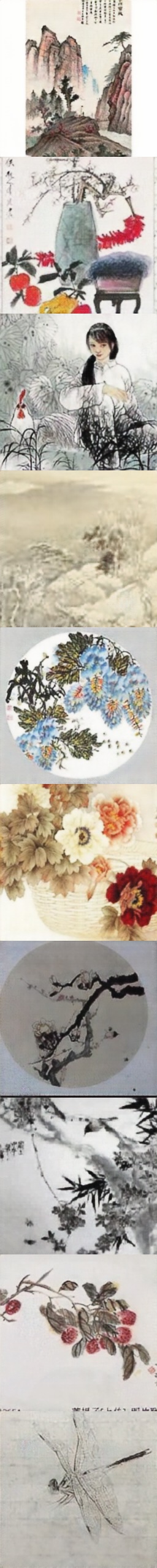}

\begin{subfigure}[]{0.13\linewidth}\caption*{$V$ (chinese)}\end{subfigure}
\begin{subfigure}[]{0.13\linewidth}\caption*{$G_B(V)$ (oil)}\end{subfigure}
\begin{subfigure}[]{0.13\linewidth}\caption*{$G_A(G_B(V))$ (chinese)}\end{subfigure}
\begin{subfigure}[]{0.13\linewidth}\caption*{$V$ (chinese)}\end{subfigure}
\begin{subfigure}[]{0.13\linewidth}\caption*{$G_B(V)$ (oil)}\end{subfigure}
\begin{subfigure}[]{0.13\linewidth}\caption*{$G_A(G_B(V))$ (chinese)}\end{subfigure}
\caption{Chinese paintings$\rightarrow$oil paintings translation results by DualGAN} 
\label{fig:ch2oi}
\end{center}
\end{figure*}

\begin{figure*}
\begin{center}
\includegraphics[width=0.13\linewidth]{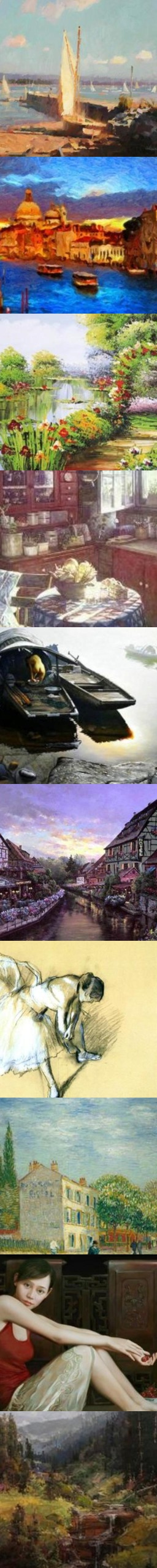}
\includegraphics[width=0.13\linewidth]{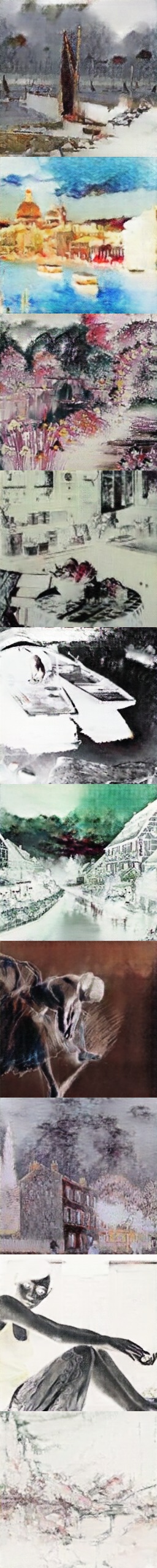}
\includegraphics[width=0.13\linewidth]{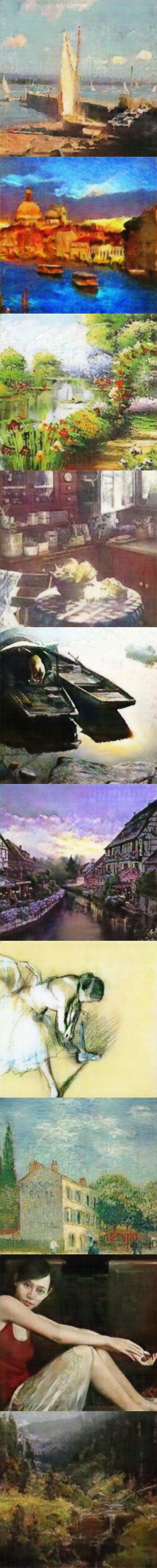}
\includegraphics[width=0.13\linewidth]{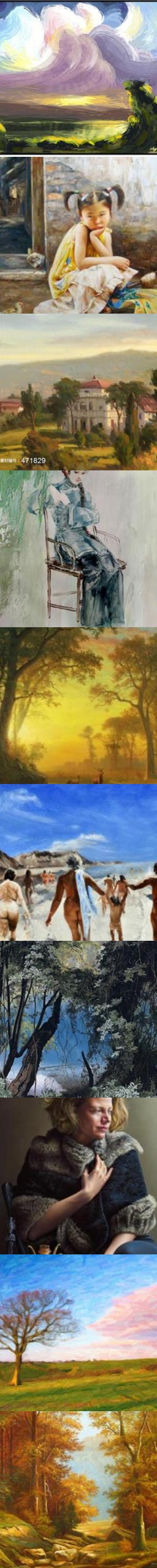}
\includegraphics[width=0.13\linewidth]{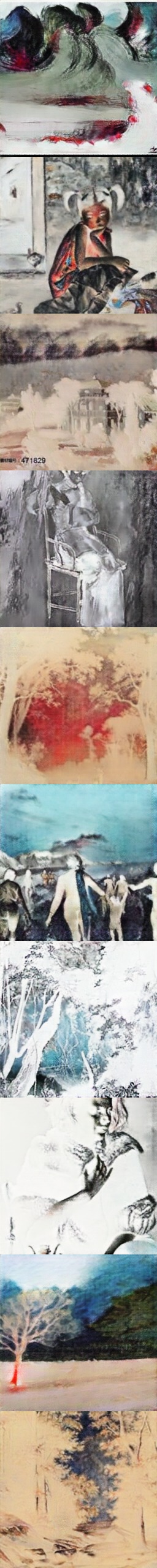}
\includegraphics[width=0.13\linewidth]{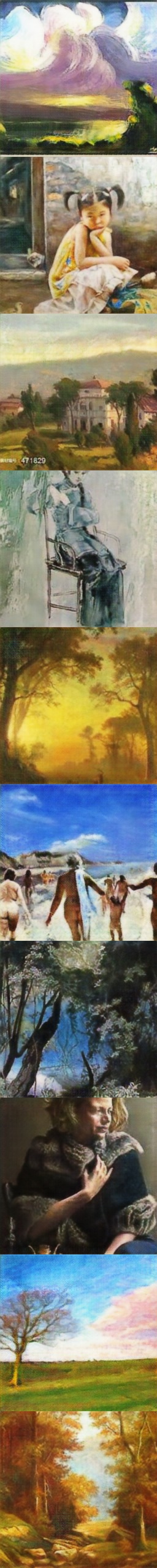}

\begin{subfigure}[]{0.13\linewidth}\caption*{$U$ (oil)}\end{subfigure}
\begin{subfigure}[]{0.13\linewidth}\caption*{$G_A(U)$ (Chinese)}\end{subfigure}
\begin{subfigure}[]{0.13\linewidth}\caption*{$G_B(G_A(U))$ (oil)}\end{subfigure}
\begin{subfigure}[]{0.13\linewidth}\caption*{$U$ (oil)}\end{subfigure}
\begin{subfigure}[]{0.13\linewidth}\caption*{$G_A(U)$ (chinese)}\end{subfigure}
\begin{subfigure}[]{0.13\linewidth}\caption*{$G_B(G_A(U))$ (oil)}\end{subfigure}
\caption{Oil painting$\rightarrow$Chinese painting translation results by DualGAN} 
\label{fig:oi2ch}
\end{center}
\end{figure*}
\section*{Appendix}
More results could be found in Figures~\ref{fig:da2ni}, \ref{fig:ph2la}, \ref{fig:sk2ph}, \ref{fig:ph2sk}, \ref{fig:la2ph}, \ref{fig:ch2oi}, \ref{fig:oi2ch}. Source codes of DualGAN have been release on \href{https://github.com/duxingren14/DualGAN}{duxingren14/DualGAN} on github.

\end{document}